\documentclass[lettersize,journal]{IEEEtran}

\IEEEoverridecommandlockouts
\usepackage{amsmath,amsfonts,amssymb, amsthm}
\usepackage{mathtools}
\usepackage{siunitx}
\DeclareMathOperator{\Tr}{Tr}
\DeclareMathOperator*{\argmax}{arg\,max}

\DeclareMathOperator*{\IS}{IS}
\DeclareMathOperator*{\simu}{sim}
\DeclareMathOperator*{\real}{real}

\newcommand{\queries}{\bar\Theta}
\newcommand{\queriesvirtual}{\hat\Theta}
\newcommand{\queriesext}{\bar X}

\newcommand{\obj}{f} %Objective
\newcommand{\Obj}{F} %Objective surrogate

\newtheorem{theorem}{Theorem}
\newtheorem{assumption}{Assumption}

\newtheorem{lemma}{Lemma}
\newtheorem{remark}{Remark}
\usepackage{algorithm}
\usepackage{algpseudocode}
\usepackage{booktabs} 
\usepackage{array, multirow}
\usepackage{bm}
\usepackage{textcomp}
\usepackage{stfloats}
\usepackage{url}
\usepackage{verbatim}
\usepackage{graphicx}
\usepackage{balance}
\usepackage{subfiles}
\usepackage{subcaption}
\usepackage{tikz}
\usetikzlibrary{arrows.meta,positioning,fit,calc}
\usetikzlibrary{shapes.geometric, arrows, positioning}

\usepackage{pgfplots}
\pgfplotsset{compat=1.16}

\DeclareMathOperator*{\E}{\mathbb{E}}

\newcommand{\ie}{i\/.\/e\/.,\/~}
\newcommand{\eg}{e\/.\/g\/.,\/~}
\newcommand{\cf}{cf\/.\/~}
\newcommand{\fig}{Fig\/.\/\,}

\newcommand{\tabl}{Tab\/.\/~}

\newcommand{\theo}{Theorem~}
\newcommand{\etal}{\textit{et al.}}
\newcommand*{\tran}{^{\mkern-1.5mu\mathsf{T}}}
\definecolor{RWTHBlue}{rgb}{0, 0.32941176470588235, 0.6235294117647059}
\definecolor{real}{rgb}{0.32941176470, 0.4470588235294118, 0.6705882352941176}
\definecolor{sim}{rgb}{0.8196078431372549, 0.5333333333333333, 0.33725490196078434}
\newcommand{\direction}{\nu}

\newcolumntype{C}[1]{>{\centering\arraybackslash$}m{#1}<{$}}
\newlength{\colwd}                    
\algnewcommand{\algorithmicgo}{\textbf{go to}}
\algnewcommand{\Goto}[1]{\algorithmicgo~\ref{#1}}

\usepackage[isbn=false,
 backend=biber,
 style=ieee,
 bibstyle=ieee, % write literature lists in IEEE style
 citestyle=numeric-comp, % \cite uses a numeric key
 sortcites=true, 
 giveninits=true, % use caps 
 backref=false]{biblatex}

\DeclareRobustCommand{\citet}[1]{\citeauthor{#1}~\cite{#1}}

\usepackage{hyperref}
\hypersetup{
    colorlinks   = true, % Colors links instead of ugly boxes
    urlcolor     = blue, % Color for external hyperlinks
    linkcolor    = black, % Color for internal links (e.g., sections, references)
    citecolor    = blue, % Color for citations
    pdftitle     = {Simulation-Aided Policy Tuning for Black-Box Robot Learning}, % Title for the PDF
    pdfauthor    = {Shiming He, Alexander von Rohr, Dominik Baumann, Ji Xiang, Sebastian Trimpe}, % Author for the PDF metadata
}

\tikzset{
  frame/.style={
    rectangle, draw,
    text width=6em, text centered,
    minimum height=3em,fill=RWTHBlue!10,
    rounded corners,
  },
  framebig/.style={
    rectangle, draw,
    text width=10em, text centered,
    minimum height=5em,fill=RWTHBlue!10,
    rounded corners,
  },
  frametransparent/.style={
    rectangle, draw,
    rounded corners,
  },
  line/.style={
    draw, -{Latex},rounded corners=3mm,
  },
}

\begin{document}

\title{Simulation-Aided Policy Tuning\\ for Black-Box Robot Learning} 

\author{Shiming He, Alexander von Rohr, Dominik Baumann, Ji Xiang, Sebastian Trimpe% <-this % stops a space
\thanks{Simulations were performed with computing resources granted by RWTH Aachen University under project RWTH1205. (Corresponding author: Alexander von Rohr)}%
\thanks{Shiming He is with the School of Information and Electrical Engineering, Hangzhou City University, Hangzhou 310015, China (e-mail: hsm@hzcu.edu.cn)}
\thanks{Alexander von Rohr and Sebastian Trimpe are with the Institute for Data Science in Mechanical Engineering, RWTH Aachen University, Germany (e-mail: vonrohr@dsme.rwth-aachen.de, trimpe@dsme.rwth-aachen.de)}%
\thanks{Dominik Baumann is with the Department of Electrical Engineering and Automation, Aalto University, Espoo, Finland and the Department of Information Technology, Uppsala University, Uppsala, Sweden (e-mail: dominik.baumann@aalto.fi)}%
\thanks{Ji Xiang is with the College of Electrical Engineering, Zhejiang University, Hangzhou 310027, China and the Huzhou Institute of Zhejiang University, Huzhou 313000, China. (e-mail: jxiang@zju.edu.cn)}%
}

\markboth{}{}

\maketitle

\begin{abstract}
How can robots learn and adapt to new tasks and situations with little data? Systematic exploration and simulation are crucial tools for efficient robot learning. We present a novel black-box policy search algorithm focused on data-efficient policy improvements. The algorithm learns directly on the robot and treats simulation as an additional information source to speed up the learning process. At the core of the algorithm, a probabilistic model learns the dependence of the policy parameters and the robot learning objective not only by performing experiments on the robot, but also by leveraging data from a simulator. This substantially reduces interaction time with the robot. Using this model, we can guarantee improvements with high probability for each policy update, thereby facilitating fast, goal-oriented learning. We evaluate our algorithm on simulated fine-tuning tasks and demonstrate the data-efficiency of the proposed dual-information source optimization algorithm. In a real robot learning experiment, we show fast and successful task learning on a robot manipulator with the aid of an imperfect simulator.
\end{abstract}


\newcommand{\figsimtoreal}{
    \begin{figure}[!t]
        \centering       \includegraphics[width=3.5in]{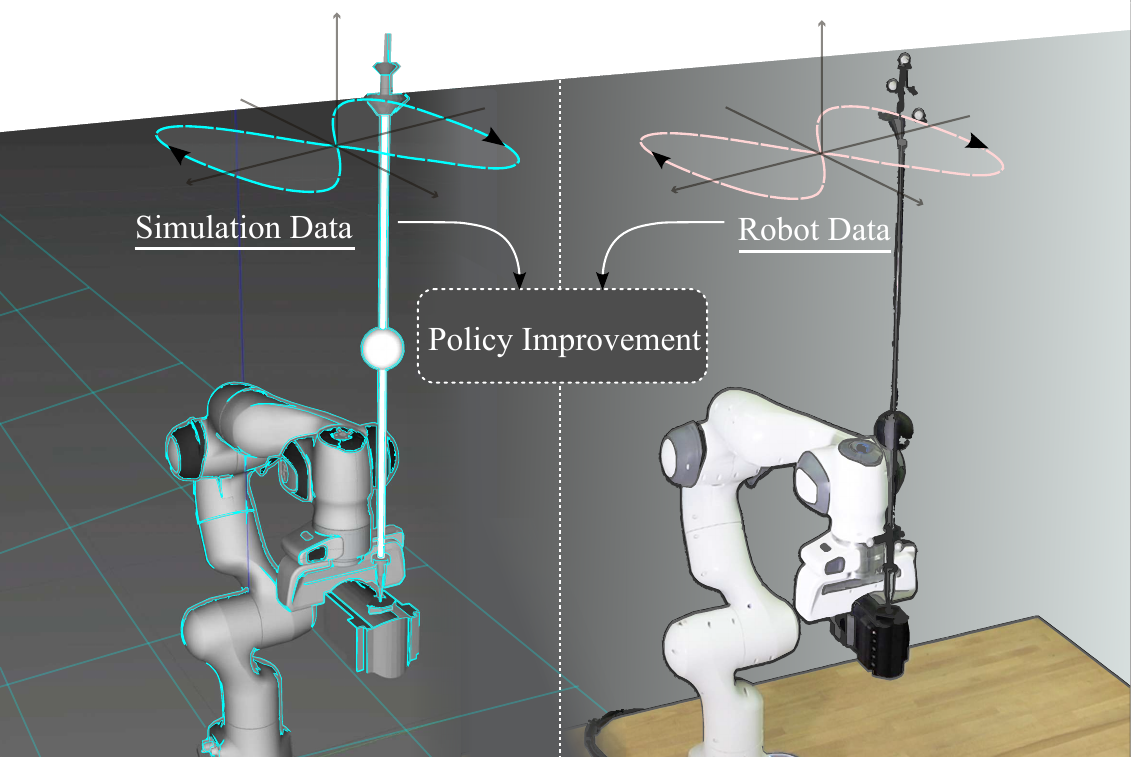}
        \caption{\textbf{Experiment setup:}  A robot manipulator is balancing a planar pendulum and learning to follow the reference trajectory with the pendulum.}
        \label{fig:sim-and-real}
    \end{figure}
} 

\section{Introduction}\label{sec:intro}

\IEEEPARstart{R}{obotic} systems need to adapt autonomously to handle their assigned tasks effectively when they are to operate in unstructured and changing environments.
However, analytical methods for the design of fine-tuned robot controllers require significant expertise to design and test.
Data-driven methods, where robots learn policies and behaviors directly by interacting with their environment, offer an attractive alternative.
Learning can enable robots to adapt to complex and dynamic environments, making them more versatile and useful in a wide range of applications.
Indeed, over the last decade, machine learning methods, particularly deep reinforcement learning (RL), have enabled robots to solve tasks in difficult and unstructured environments \cite{levine2018learning,openai2019solving,lee2020learning}.
Most RL approaches are formulated around state-action pairs and assume a Markovian environment, requiring direct state transitions and actions for their policy updates.
Further, the use of uncorrelated random noise in the action space for exploration can lead to jerky motions \cite{ibarz2021train}, which are undesirable in real-world settings, since actuators may take a lot of wear and tear during the training.

An alternative to address the above limitations is black-box policy search~\cite{chatzilygeroudis2019survey}.
Black-box policy search directly learns a mapping between the policy parameters and the expected return.
This alleviates the need to explore directly in the action space and allows tuning episodic policies, such as movement primitives (MPs) \cite{ijspeert2002movement, paraschos2013probabilistic}, which lead to smooth trajectories.
Further, it does not require Markov assumptions on states, actions, and rewards. We can, \eg define a reward that depends on time and only consider the final state.
This is suitable for robot applications, \eg pick-and-place.
The experimental setup for step-based RL on hardware still requires a specialized setup \cite{luo2024serl}. 
In black-box policy search, the robot experiment can be performed independently of the algorithm since we only need the final trajectory to evaluate the performance.
Black-box approaches, however, do not collect and use all state variables, and do not use all available information for tuning \cite{chatzilygeroudis2017black}.

\figsimtoreal

Data-efficient black-box robot learning seeks to develop algorithms that can learn effective control policies with limited \emph{interaction time} with the robot.
Here, Bayesian optimization (BO), a black-box optimization method, has successfully been applied to this problem; however, it is typically only suitable for low dimensional problems \cite{chatzilygeroudis2019survey}. How to apply BO to high-dimensional search spaces is a popular research topic (\eg\cite{eriksson2019scalable, muller2021local, frohlich2021cautious, eriksson2021high, nguyen2022local, ziomek2023random}).
Departing from the standard BO formulation and its limitations, but building on some of its key ideas, we propose a policy search and tuning method that is both data-efficient and suitable for higher dimensional black-box learning problems as it systematically leverages data from a simulator to decrease hardware interaction time.

For many robotic systems, simulators are available that yield data at a comparatively much lower cost than real robotic systems in terms of running time, mechanical wear, consumed energy, etc.
But, since models are imperfect, the learned behavior will be sub-optimal when applied to the real system.
This motivates learning, at least in part, on the robot directly.
To determine the trade-off between simulation and robot data, we propose to model the uncertainty of the simulated interaction and use Bayesian decision-making to decide on the information source (see \fig\ref{fig:sim-and-real}).
Specifically, the learner shall guarantee local policy improvement and iteratively decide which information source should be used to provide the desired guarantees. 
If the simulation data is informative, we avoid data collection in the physical world, decreasing the overall cost of data acquisition. However, once the information from the simulation is insufficient to further improve the policy on the current task, data from the robot will be needed.
Following a policy change, the simulator might once more prove to be a valuable information source.
This interactive approach differs from existing methods that leverage simulators and tackle the problem of the so-called reality gap using, \eg domain randomization~\cite{james2017transferring, peng2018sim, andrychowicz2020learning} or domain adaption~\cite{gupta2017learning, rusu2017sim, tanwani2021dirl},
as in our method the learner actively decides which data source it needs to improve the policy for the robot (not the simulation) during the learning process.

\paragraph*{Contributions}
We propose a general and data-efficient zero-order local search method suitable for black-box learning problems often encountered in robot applications.
The method, based on \citet{muller2021local}, employs zero-order and multi-fidelity information, where a probabilistic model directly learns the dependence between policy parameters and the robot learning objective.
In summary, our contributions are:
\begin{itemize}
    %\item A general and data-efficient zero-order local search method that allows for parameterized policies with unknown gradients. 
   % \item A multi-fidelity local BO method that can use a simulator in the learning process as an imperfect (\ie biased) information source to reduce the data needed from the real robotic system.
    \item An efficient decision rule for policy updates with multi-fidelity local BO that ensures a high probability of improvement at every iteration.
    \item Evaluations of the suggested local search method on both synthetic benchmarks and robotic tasks.
\end{itemize}

Our method is especially well-suited to fine-tuning tasks or deployment in dynamic environments where fast adaptation is required and robots must learn and adapt with minimal interaction time. The dual-information source optimization, which leverages both real-world data and simulations, enables fast improvements on hardware through guided exploration and simulations.

\newcommand{\figblock}{
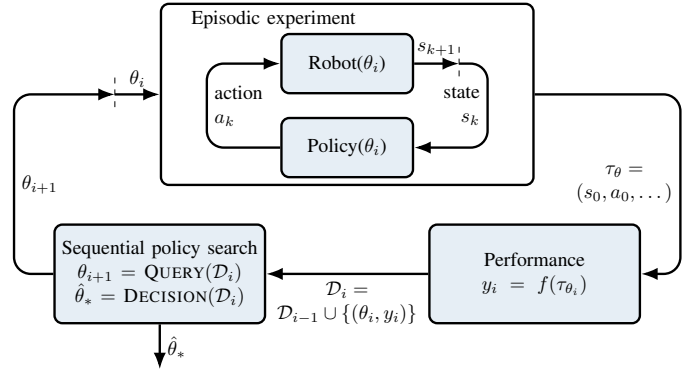
\begin{figure}
\centering
\resizebox{3.5in}{!}{\begin{tikzpicture}[very thick,node distance = 4cm]

\node [frame] (environment) {Robot($\theta_i$)};
\node [frame, below=0.4 of environment] (agent) {Policy($\theta_i$)};

\draw[line] (agent) -- ++ (-2.5,0) |- (environment) node[right,pos=0.25,align=left] {action\\ $a_k$};

\coordinate[right=0.8 of environment] (P);
\coordinate[above=0.3 of P] (Pa);
\coordinate[below=0.3 of P] (Pb);

\draw[thin,dashed] (Pa) |- (Pb);

\draw[line] (environment.east) -- (P) node[midway,above]{$s_{k+1}$};

\draw[line] (environment.east -| P) -- (2.5,0) |- 
(agent.east) node[left, pos=0.25, align=right] {state\\ $s_k$};

\coordinate[left=3.2 of environment.north] (LUh);
\coordinate[above=.4 of LUh] (LU);
\coordinate[right=3.2 of agent.south] (RL);

\node[frametransparent, fit=(agent) (environment) (LU) (RL)] (system) {};
\node[below] at (system.north) {Episodic experiment \hspace*{2.5cm}};

\node [framebig, below = 2.3 of system.east] (cost) {Performance \\ $y_i = f(\tau_{\theta_i})$};

\node [framebig, below = 2.3 of system.west] (acquisition) 
{
Sequential policy search \\
$\theta_{i+1} = \textsc{Query}(\mathcal{D}_{i})$ \\
$\hat \theta_* = \textsc{Decision}(\mathcal{D}_{i})$
};

\coordinate[below=0.8 of acquisition] (out);

\draw[line] (acquisition.south) -- (out) node[midway,right]{$\hat \theta_*$};

\draw[line] (cost) -- (acquisition) node[below,pos=0.5,align=center] {
$\mathcal{D}_{i} =$ \\ $\mathcal{D}_{i-1} \cup \{(\theta_i, y_i)\}$
};

\coordinate[left=0.8 of system] (Q);
\coordinate[above=0.3 of Q] (Qa);
\coordinate[below=0.3 of Q] (Qb);

\draw[thin,dashed] (Qa) |- (Qb);
\draw[line] (Q) -- (system.west) node[midway,above]{$\theta_{i}$};

\coordinate[left=0.7 of acquisition] (sys_in);
\coordinate[right=0.7 of cost] (sys_out);

% (environment.east -| P)

\draw[line] (system.east) -| (sys_out) node[left, pos=0.75,align=center] {
$\tau_\theta =$ \\ $(s_0, a_0, \dots )$
} -- (cost.east);

\draw[line] (acquisition.west) -- (sys_in) |- (Q) node[right, pos=0.25, align=right] {$\theta_{i+1}$};

\end{tikzpicture}}
\caption{\textbf{Sequential black-box policy search:} The search algorithm determines a query in order to gain more information on the performance function. The policy is then evaluated with the parameters given by the query. In each iteration, the search algorithm decides on its current guess of the best policy.}
\label{fig:block_diagram}
\end{figure}
}

\section{Problem formulation}\label{sec:problem}

\figblock
We formulate robot learning as an episodic policy search problem to optimize closed-loop trajectories $\tau = (\bm s_0, a_0, \dots, \bm s_{K-1}, a_{K-1})$ of $K$ states $s$ and actions $a$ (\cf\fig\ref{fig:block_diagram}). 
The closed-loop system is parameterized through a vector $\theta \in \Theta$ that lies in a bounded and closed $\Theta \subset \mathbb{R}^d$.
The parameters $\theta$ represent any configurable property of the robotic system. Specifically, in the experimental sections of this paper we use the parameters to define a MP. We denote a trajectory that was obtained with a specific configuration $\theta$ as $\tau_\theta$.
We define the learning task in terms of a performance function $\obj : (\mathcal{S} \times \mathcal{A})^K \mapsto \mathbb{R}$ that maps a trajectory of the closed-loop to a scalar.
The episodic policy search problem is cast as a maximization problem over $\obj$,
\begin{equation}\label{eq:min_cost}
    \theta^* = \!\argmax\limits_{\theta \in \Theta}  \obj(\tau_\theta).
\end{equation}
For notational convenience we use $\obj(\tau_\theta)$ and $\obj(\theta)$ interchangeably.

We solve the policy search problem by sampling trajectories $\tau$ with different policy parameterizations (\cf\fig\ref{fig:block_diagram}), where each sample obtained through an experiment.
After the \(i\)-th experiment, we receive a noisy performance sample 
\begin{equation}
    y_i = \obj(\tau_{\theta_i}) + \epsilon_i, 
\end{equation}
where we assume $\epsilon \sim \mathcal{N}(0,\sigma^2_n)$.
The reason for noisy evaluations are, for example, stochastic dynamics or random initial conditions.
In contrast to classical optimization and RL, we consider the policy search problem as a \emph{black-box optimization} problem and only work with tuples $(\theta, y)$. 
Importantly, we do not require an analytical expression of $\obj$ or any first-order information or samples, \ie $\nabla \obj$ is unknown, and we do not require assumptions on the level of the underlying dynamical system or policies.

In addition to trajectories from the robotic system, the algorithm can acquire additional data from an imperfect simulator
\begin{equation}\label{eq:sim}
    \obj_\mathrm{sim}(\theta) = \obj(\theta) + \obj_\mathrm{gap}(\theta),
\end{equation}
with an unknown reality gap $\obj_\mathrm{gap}$.
The performance on the simulator will differ from that on the robot. This difference constitutes a bias and data from a policy evaluated with \eqref{eq:sim} leads to a biased estimate $\obj_\mathrm{sim}(\theta)$ of $\obj(\theta)$.

To build a probabilistic surrogate model for Bayesian decision making, we assume some prior knowledge of the functions $\obj$, $\obj_\mathrm{sim}$, and their correlations.
Specifically, we assume $\obj$ and the reality gap $\obj_\mathrm{gap} = \obj - \obj_\mathrm{sim}$ are samples from known Gaussian process (GP) priors.
Since GPs are closed under addition, $\obj_\mathrm{sim}$ is also a GP.
This assumption that the objective functions are samples from a GP is standard in BO (\cf \cite{srinivas2010gaussian}).
\begin{assumption}
\label{ass:lipschitz}
The performance functions $\obj$ and $\obj_\mathrm{gap}$ are samples from a known robot GP prior \(\Obj \sim \mathcal{GP} (m, k)\) and a gap GP prior \(\Obj_\mathrm{gap} \sim \mathcal{GP} (m_\mathrm{gap}, k_\mathrm{gap})\), respectively, whose mean functions $m : \Theta \mapsto \mathbb{R}$ are at least once differentiable and whose covariance functions $k : \Theta \times \Theta \mapsto \mathbb{R}$ are at least twice differentiable. 
Further, the derivatives of $\obj$ are Lipschitz continuous with constant \(L > 0\), i.e., we have that \( \|\nabla \obj(a) - \nabla \obj(b)\| \leq L\|a - b\| \) for any \(a, b\).
\end{assumption}

\section{Related Work}
Our work provides a method for data-efficient robot learning that builds on Bayesian decision-making, as well as learning through simulation.
This section provides an overview of related work in these areas.

\paragraph*{Data-efficient robot learning}
There are currently two main strategies for improving data-efficiency in robot learning (\cf \cite{chatzilygeroudis2019survey}). 
One is to build prior knowledge, \eg of the policy representation, into a learning system.
% The prior knowledge can exist on both policy representation and expected return.
The other is to learn the robot dynamics and optimize the policy w.r.t.\ the learned model. This strategy is known as model-based policy search.

Prior knowledge for policy representation is often motivated by control theory, where human experts decide which control design is suitable for the given task~\cite{berkenkamp2016safe,marco2016automatic,edwards2021automatic,guzman2022bayesian}.
The learning task then consists of tuning the parameters of the chosen policy representation and is thus frequently called ``controller tuning''.
Choosing such a parameterization requires specific knowledge of the system dynamics.
Further, prior knowledge can also be encoded into trajectory-based policies.
In this setting, the policy is designed as a sequence of way-points~\cite{jain2020computing} or uses function approximators to mimic system dynamics~\cite{solak2019learning}. 
In the latter case, MPs are the most widely used policy structure.
Located at a higher level, \ie a task level, trajectory-based policies can be generated by a human expert or by a pre-trained RL model \cite{otto2023deep}.
It is a common practice to have a low-level controller to facilitate the trajectory tracking.
In these cases, an optimizer tries to fine-tune the trajectories for the given robotic tasks.
In the robot example of this paper, we leverage these policy parameterizations and learn to improve a task trajectory encoded by MPs to achieve better task performance.

To make RL more data-efficient model-based policy search algorithms learn a model of the dynamics~\cite{deisenroth2011pilco,chatzilygeroudis2017black,chua2018deep, buckman2018sample, janner2019when, yang2020data, whitman2023learning} and optimize the policy with respect to this model. However, these algorithms are not designed for the episodic, black-box setting we consider herein and need additional assumptions on the dynamics and reward signal.

\paragraph*{Bayesian optimization for robot learning}
Bayesian optimization, \ie using Bayesian decision theory to find the optimizer of a function in a data-efficient way, has been re-discovered multiple times in various communities.
We focus here on BO for robot learning and refer the reader to \citet{garnett2023bayesian} for an in-depth introduction to BO.
In policy search for robot learning, BO uses Bayesian inference and a sequential design of experiments (DoE) to find policies in a data-efficient way.
Bayesian optimization was first applied to robot learning to solve problems in the form of \eqref{eq:min_cost} by \citet{lizotte2007automatic}, where a quadruped robot learned gaits by optimizing the parameters of a walk engine.
Since then, BO has been successfully applied to robot learning \cite{marco2016automatic, driess2017constrained, marco2017virtual, frohlich2021cautious}.
%via controller tuning \cite{marco2016automatic, driess2017constrained, marco2017virtual, guzman2022bayesian} and direct policy learning \cite{calandra2016bayesian,rai2018bayesian,jaquier2020bayesian,frohlich2021cautious,muller2021local,frohlich2021cautious,marco2021robot,yuan2019bayesian}.
These prior works show that BO is a viable tool for data-efficient robot learning, motivating our adoption of a similar approach.
However, the described methods are only applicable to policies with a relatively small number of policy parameters (\cf\cite{chatzilygeroudis2019survey})
as they search globally for the optimal policy and global searches are highly susceptible to the curse of dimensionality.
With limited interaction time global search is likely to be outperformed by a local search \cite{muller2021local,frohlich2021cautious}.

The local search method presented in this paper does not solve the global optimization problem. Instead, it searches for a local improvement of a given parameter.
In local search variants of BO the search is restricted using trust-regions \cite{eriksson2019scalable}, local search distributions \cite{akrour2017local}, or uncertainty about the objective \cite{frohlich2021cautious}.
Instead of constraining the search space to force local exploration, we reformulate the objective for the sequential design of experiments.
The new formulation is based on a DoE algorithm, first proposed in \cite{muller2021local}, that finds a search direction in the parameter space to improve the current guess of the best policy.
In contrast to BO, the data is used to optimally reduce the uncertainty of the objective's gradient at the current guess of the best policy in \cite{muller2021local} or to maximize the probability of improvement in \cite{nguyen2022local}.
We extend these results to obtain probabilistic guarantees for policy improvement.
We then adaptively choose how much data is needed for these guarantees at each policy update, thereby avoiding unnecessary interaction and further increasing the data-efficiency.

\paragraph*{Sim-to-real robot learning}
Sim-to-real robot learning is often understood as first learning a policy in simulation, and then transferring it to the robot. Unfortunately, the reality gap---the mismatch between the simulation and the real system---can impede such a transfer.
Numerous algorithms have been proposed to tackle the sim-to-real transfer problem.
One popular approach to account for the bias in the simulator is domain randomization  \cite{tobin2017domain, james2017transferring, peng2018sim, andrychowicz2020learning} (see \citet{muratore2022robot} for a survey).
Domain randomization varies the parameters of the simulator and learns a policy that is robust to these variations.
The underlying assumption is that the variations will account for the bias of the simulator. 
By contrast, we account for the bias of the simulator by assuming imperfect correlations in the expected return of a policy.
Then, instead of increasing the variance of the simulator and learning a robust policy, we model both returns probabilistically and reason about the probability of improvement on the robot.
We can then sequentially decide where to query data.
%
%Building on domain randomization domain adaption methods assumes that different domains share common characteristics so that it is possible to learn an invariant feature for transfer between simulator and real system~\cite{gupta2017learning, rusu2017sim, tanwani2021dirl}. 

Our proposed method is closely related to the work by \citet{marco2017virtual}, where a dual-information source GP model is proposed; however, the algorithm in \cite{marco2017virtual} is a global search method, and thus does not scale well with the dimensionality of the policy. 
Instead, we propose a local search where we improve the current guess of the best policy on the robot. %using a simulator. 

\section{Preliminaries}
This section introduces the necessary mathematical concepts for the proposed algorithms.
In particular, we introduce derivative GP models and an exploration strategy to reduce gradient uncertainty from \cite{muller2021local}.

\subsection{Derivative GP Model}
\label{sec:gp_model}
Assumption~\ref{ass:lipschitz} 
ensures that the GP is mean-square differentiable \cite{rasmussen2006gaussian} and therefore the posterior over the objective's derivative is also a GP.
This posterior distribution can be computed analytically, which, following \citet{muller2021local} we will use to form a belief over the function's derivative from noisy zeroth-order observations.

The joint distribution between noisy zeroth-order observations and the derivative at $\hat\theta_*$ is normally distributed
\begin{equation} \label{eq:model_GIBO}
\begin{aligned}
&\begin{bmatrix}
	y \\ \nabla F({\hat\theta_*})
\end{bmatrix} \sim \\
&\mathcal{N} \left(
\begin{bmatrix}
    \mu(\queries) \\
    \nabla \mu(\hat\theta_*)
\end{bmatrix}, 
\begin{bmatrix}
	K(\queries,\queries) + \sigma^2 I & \nabla  K(\queries, \hat\theta_*) \\
	\nabla K(\hat\theta_*, \queries) & \nabla^2  K(\hat\theta_*, \hat\theta_*)
\end{bmatrix} \right),
\end{aligned}
\end{equation}
with $n$ zeroth-order observations $y$, whose locations are $\queries = [\theta_1, \dots,\theta_n] \subset \Theta$, and \(\nabla F\) a belief over \(\nabla f\). $K$ is the covariance matrix given by the kernel function $k: \Theta \times \Theta \mapsto \mathbb{R}$.
By conditioning the prior joint distribution on the observation, we obtain the Gaussian posterior
\begin{align}\label{eq:posterior_derivative_gp}
\begin{split}
&\nabla F({\hat\theta_*}) \big| \queries, y \sim \mathcal{N}(\mu_*^\prime, \Sigma_*^\prime) \\
\mu_*^\prime =& \underbrace{\vphantom{\left(K(\queries,\queries) + \sigma^2 I \right)^{-1}}\nabla \mu(\hat\theta_*)}_{\in \mathbb{R}^{d}} \\
&+ \underbrace{\vphantom{\left(K(\queries,\queries) + \sigma^2 I \right)^{-1}}\nabla K(\hat\theta_*, \queries)}_{\in \mathbb{R}^{d\times n}} \underbrace{\left(K(\queries,\queries) + \sigma^2 I \right)^{-1}}_{\in \mathbb{R}^{n\times n}} \underbrace{\vphantom{\left(K(\queries,\queries) + \sigma^2 I \right)^{-1}} (y - \mu(\queries))}_{\in \mathbb{R}^n} \\
\Sigma_*^\prime =& \underbrace{\vphantom{\left(K(\queries,\queries) + \sigma^2 I \right)^{-1}}\nabla^2 K(\hat\theta_*,\hat\theta_*)}_{\in \mathbb{R}^{d\times d}}\\
&- \underbrace{\vphantom{\left(K(\queries,\queries) + \sigma^2 I \right)^{-1}}\nabla K(\hat\theta_*,\queries)}_{\in \mathbb{R}^{d\times n}} \underbrace{\left(K(\queries,\queries) + \sigma^2 I \right)^{-1}}_{\in\mathbb{R}^{n\times n}} \underbrace{\vphantom{\left(K(\queries,\queries) + \sigma^2 I \right)^{-1}}\nabla K(\queries,\hat\theta_*)}_{\in\mathbb{R}^{n\times d}},
\end{split}
\end{align}
with $\mu_*^\prime \in \mathbb{R}^d$ and $\Sigma_*^\prime \in \mathbb{R}^{d\times d}$.

\subsection{Sequential Policy Search with Gradient Information}
\label{sec:gradient}
Sequential black box policy search (cf.~\fig\ref{fig:block_diagram}) requires both a query function and a decision function.
We build our algorithm on Gradient Information with BO (GIBO)~\cite{muller2021local}.
This method maintains a current guess of the best policy $\hat \theta_*$ and queries the objective to find a search direction that improves this guess.
A GIBO query is defined as the observation that optimally reduces a measure of the total uncertainty of the gradient for the current guess of the best policy $\Sigma_*^\prime$
\begin{equation}
\begin{aligned}\label{eq:gi_query}
\textsc{Query}(\mathcal{D},\hat \theta_*) \coloneqq \argmax_{\theta \in \Theta} \alpha_\text{GI}(\theta\mid \mathcal{D}, \hat \theta_*), 
\end{aligned}
\end{equation}
where $\alpha_\text{GI}$ is the Gradient Information (GI) acquisition function
\begin{equation}
\begin{aligned}\label{eq:gi_acquistion}
& \alpha_\text{GI}(\theta\mid\mathcal{D}, \hat \theta_*) \\
= &\E{[
\Tr(\Sigma_*^\prime\mid\mathcal{D}) -
\Tr(\Sigma_*^\prime\mid\mathcal{D} \cup (\theta, y)) ]} \\
= & \Tr \left(\nabla K\left(\hat \theta_*,\queriesvirtual\right) \left(K\left(\queriesvirtual,\queriesvirtual\right) + \sigma_n^2 I \right)^{-1} 
\right.\\& \qquad\qquad\qquad\left. 
\nabla K(\hat\theta_*, \queriesvirtual)\right)\tran,
\end{aligned}
\end{equation}
with $\Sigma_*^\prime$ the gradient variance \emph{before} and \emph{after} observing a new point $(\theta, y)$, $\queriesvirtual = \queries \cup {\theta}$.
Critically, the posterior variance does not depend on the observation $y$. Hence, the optimal query can be determined without knowing its outcome.
Here, the trace is taken as a measure of total variance, but other measures are possible.
For a full derivation, see \cite{muller2021local}.
The original GIBO algorithm now makes $M$ such queries, where $M$ is set to be the dimensions of the policy domain, before updating $\hat\theta_*$ using the mean estimate of the gradient
\begin{equation}\label{eq:update_rule}
    \hat\theta_{*} \leftarrow \hat\theta_{*} - \eta \mu_{*}^\prime,
\end{equation}
with step size $\eta$.
Increasing $M$ could slow down the optimization since using a larger number of samples does not necessarily provide any additional valuable gradient information when an accurate estimation has been achieved.
On the other hand, if $M$ is reduced too much, while it might help avoid unnecessary queries and improve efficiency, there is a risk of erroneous gradient estimates.
A core contribution of our work is to replace the hyperparameter $M$ with a condition on the probability of improvement, meaning we query as long as necessary to guarantee an improved policy with high probability.

\section{High-Confidence Policy Improvements with Gradient Information Bayesian Optimization}

In this section, we introduce the Gradient Information with BO (GIBO) framework \cite{muller2021local}.
We then propose our own algorithm High-Confidence Improvement GIBO (HCI-GIBO), which introduces an efficient decision rule for when to update the current guess of the best policy $\hat \theta_*$.
Furthermore, we extend the method such that it can leverage data from an additional biased information source such as a simulator (S-HCI-GIBO).

\begin{algorithm}[t]
\small
\caption{GIBO (adapted from \cite{muller2021local})}\label{algo:gibo}
\begin{algorithmic}[1]
    \State \textbf{Hyperparameters}: hyperpriors for GP hyperparameters,
    improvement confidence $\alpha$,
    %SimToReal switching threshold $T$,
    stepsize $\eta$,
    \State $\hat{\theta}_* \leftarrow \theta_0$, $\mathcal{D}  \leftarrow \mathcal{D}_0$
    \Repeat \Comment{Policy updates}
    \Repeat
    \Comment{DoE for descent direction}
    \State $\theta_i \leftarrow \textsc{Query}(\mathcal{D}, \hat\theta_*)$
    \State $y_i \leftarrow f(\theta_i) + \epsilon_i$
    \State $\mathcal{D}  \leftarrow \mathcal{D} \cup (\theta_i, y_i)$
    \Until{$\textsc{Commitment}(\mathcal{D}, \hat\theta_*)\geq \alpha$}
    \State $\nabla f \leftarrow \textsc{DescentDirection}(\mathcal{D}, \hat\theta_*)$ 
    \State $\hat\theta_* \leftarrow \hat\theta_* - \eta \, \nabla f$ \Comment{Optional: normalized gradient descent}
    \Until{$\textsc{Termination}(\mathcal{D}, \hat\theta_*)$}  
\end{algorithmic}
\end{algorithm}
\subsection{Gradient Information with BO}
The general GIBO framework is summarized in Alg.~\ref{algo:gibo}.
Just like BO, GIBO requires hyperpriors for GP hyperparameters.
For the SE kernel, GP hyperparameters are lengthscales that describe how relevant an input is.
In practice, hyperpriors, \ie prior distributions of hyperparameters, are determined through domain knowledge. 
A common approach to obtain such experience is to collect a small dataset before the optimization.
During the optimization, hyperparameters (lengthscales) can be adjusted by maximum a posteriori estimation (see chapter 5 of \cite{rasmussen2006gaussian} for details).

The algorithm relies on three decisions, namely, a \textsc{Query} function that decides where to query the objective function next; a \textsc{Commitment} condition that decides when to terminate the query process; and a \textsc{DescentDirection} function that guides the gradient descent to improve the policy. These three elements, which we will make precise for HCI-GIBO, operate in two nested loops. In the inner loop, we query the objective function. Once the commitment condition is triggered, we leave the inner loop and take a gradient step with the learned descent direction.

For our work, the derivative GP model (Sec.~\ref{sec:gp_model}) serves as a proxy to approximate the unknown objective and to derive an analytical distribution for the objective’s derivative.
Then, we use the gradient information introduced in Sec.~\ref{sec:gradient} as the \textsc{Query} function, to actively locate the most informative query points for the gradient of the current guess of the best policy.
Compared with GIBO, we reduce data redundancies by deriving a criterion to quantify the improvement confidence and to terminate the query process. This improvement confidence serves as \textsc{Commitment} in HCI-GIBO. %and is introduced in the following section.

\subsection{High-Confidence Improvement}

The Bayesian design of experiments to infer the gradients in GIBO algorithms contributes to data efficiency by actively querying data to improve the estimates. Intuitively, more data leads to a better gradient estimate. But how much data or knowledge does the optimizer require to improve the policy?
In previous work \cite{muller2021local,nguyen2022local}, the number of queries is a hyperparameter\footnote{\citet{nguyen2022local} use a fixed number of queries, but make multiple steps.}.
By contrast, HCI-GIBO queries the function until it can guarantee that the update step will improve the function with high probability. 
This avoids early updates that might lead to a decrease in performance and the collection of unnecessary data.

To guarantee improvement, we first define the set of all vectors that improve the objective.
\begin{lemma}\label{thm:deterministic_improvement}
(Deterministic Improvement) Assume the cost functional \(f\) is differentiable and that its gradient is Lipschitz continuous with constant \(L > 0\). Let \(\nabla f\) be the gradient of \(f\) and \(\direction\) the descent direction.
Then, for all \(\direction\) and a given fixed step size $\eta$, the function improves, \ie
\begin{equation}
\label{eqn:deterministic_improvement_claim}
    f(\theta - \eta\direction) < f(\theta)   
\end{equation}
if 
\begin{equation} \label{eq:criteria}
    \left\langle \frac{\direction}{\|\direction\|}, \nabla f \right\rangle > \frac{L }{2} \eta \|\direction\|.
\end{equation}
\end{lemma}
\begin{proof}
Let \(t\) be a scalar parameter and define \(r(t) \coloneqq f(\theta-t\eta \direction)\). The derivative of \(r(t)\) with respect to \(t\) is
\begin{equation}
    \frac{{\rm d}r}{{\rm d}t} = -\eta  \nabla f(\theta-t\eta \direction) \tran \direction.
\end{equation}
Then,
\begin{equation}
\begin{aligned}
    &f(\theta - \eta \direction) - f(\theta) = r(1) - r(0) = \int_0^1 \frac{{\rm d}r}{{\rm d}t} {\rm d}t\\
    &= \int_0^1 -\eta \nabla f\left(\theta-t\eta \direction\right)\tran \direction {\rm d}t\\
    &\leq \int_0^1 -\eta \nabla f(\theta)\tran \direction {\rm d}t + \eta\left|\int_0^1 \left[\nabla f(\theta-t\eta \direction) - \nabla f(\theta)\right]\tran \direction {\rm d}t\right|\\
    &\leq \int_0^1 -\eta \nabla f(\theta) \tran \direction {\rm d}t + \eta \|\direction\| \int_0^1 \|\nabla f(\theta-t\eta \direction)-\nabla f(\theta)\| {\rm d}t.\\
\end{aligned}
\end{equation}
Since $\nabla f$ is Lipschitz continuous, we have
\begin{equation}
    \|\nabla f(\theta-t\eta \direction)-\nabla f(\theta)\| \leq L \|t\eta \direction\|.
\end{equation}
Thus,
\begin{equation}
\begin{aligned}
f(\theta - \eta \direction) - f(\theta) 
    & \leq  \int_0^1 -\eta \nabla f(\theta)\tran  \direction {\rm d}t + \eta\|\direction\|\int_0^1 L\|t\eta \direction\| {\rm d}t \\
    & = -\eta \nabla f(\theta) \tran \direction + \frac{L}{2}\eta^2\| \direction\|^2.
\end{aligned}
\end{equation}
For~\eqref{eqn:deterministic_improvement_claim} to hold, we require
\begin{align}\label{eq:19}
     -\eta \, \nabla f(\theta)\tran \, \direction + \frac{L}{2} \eta^2 \| \direction \|^2 & < 0.
\end{align}
Now \eqref{eq:19} holds by assumption \eqref{eq:criteria}, since
\begin{align}
     -\eta \, \nabla f(\theta)\tran \, \direction + \frac{L}{2} \eta^2 \| \direction \|^2 & < 0 \\
     \iff \left\langle \frac{\direction}{\|\direction\|}, \nabla f \right\rangle > \frac{L }{2} \eta \|\direction\|.
\end{align}
\end{proof}
For a given $\nabla f(\theta_*)$, Lemma~\ref{thm:deterministic_improvement} can be used to determine the set of vectors $\direction$ that guarantee improvement.
%In \fig\ref{fig:confidence}, the boundaries of this set are shown as an example.
For $\direction = \nabla f(\theta)$, we recover the well known convergence condition on the step size $\eta$ as $\eta \leq \frac{2}{L}$ \cite[Sect.~1.4.2]{polyak1987optimization}.
% \begin{equation}
%     \eta \leq \frac{2}{L}. %Boris T. Polyak - Introduction to Optimization, Page 21),
% \end{equation}
We now use Lemma~\ref{thm:deterministic_improvement} to derive a high confidence criteria to be used when $\nabla f(\theta_*)$ is unknown and replaced by a belief $\nabla F(\theta_*)$.

\begin{theorem}
\label{thm:improvement_confidence}
(Improvement Confidence) Assume the belief about the gradient at \(\theta\) is \(\nabla F(\theta) \sim \mathcal{N}(\mu, \Sigma)\).
Then, %using the mean estimate improves $F$ with probability at least $\alpha$, 
\begin{equation}\label{eq:prob_improve}
    \Pr\left[F(\theta - \eta \mu) < F(\theta)\right] \geq \alpha,
\end{equation}
if
\begin{equation} \label{eq:cdf}
\begin{aligned}
    &\Pr\left[
    \left\langle \frac{\mu}{\|\mu\|}, \nabla F(\theta)\right\rangle > \frac{L}{2} \eta \|\mu\|
    \right] \geq \alpha.
\end{aligned}
\end{equation}
\end{theorem}
\begin{proof}
The proof follows directly from Lemma~\ref{thm:deterministic_improvement}. All vectors that satisfy \eqref{eq:criteria} improve $f$. Therefore, if a realization of the random variable $\nabla F$ satisfies \eqref{eq:criteria} with at least probability $\alpha$, \eqref{eq:prob_improve} follows.
\end{proof}
We can use \theo\ref{thm:improvement_confidence} to guarantee probabilistic improvement.
We define the $\textsc{Commitment}$ function for HCI-GIBO as the left side of \eqref{eq:cdf}, \ie the probability measure, and thereby only use as many queries as are necessary to guarantee improvement with the desired probability $\alpha$.
The distribution over the scalar product is given as
\begin{equation} \label{eq:dist}
   \left\langle \frac{\mu^\prime_*}{\|\mu^\prime_*\|}, \nabla F(\hat\theta_*)\right\rangle \sim \mathcal{N}\left(\|\mu^\prime_*\|, \frac{\mu^\prime_*}{\|\mu^\prime_*\|}\tran \Sigma \frac{\mu^\prime_*}{\|\mu^\prime_*\|}\right),
\end{equation}
for which the cumulative distribution is readily computable.

The $\textsc{Query}$ function for HCI-GIBO is defined as \eqref{eq:gi_query}, as in \cite{muller2021local}, and the descent direction is defined by means of the gradient posterior distribution $\textsc{DescentDirection}(\mathcal{D}, \hat\theta_*) \coloneqq\mu_*^\prime$ (see \eqref{eq:posterior_derivative_gp} for the closed-form expression). 
\begin{remark}
It is common to use a normalized gradient step to ameliorate some of the ``slow crawling'' problem \cite{watt2020machine}. We follow such an approach in our experiments and hence the inequality \eqref{eq:cdf} can be updated by using the normalized descent direction \(\mu/\|\mu\|\).
%\begin{equation}
%     \Pr\left[\left\langle \frac{\mu}{\|\mu\|}, \nabla %\end{equation}
\end{remark}

\subsection{High-Confidence Improvement with Simulators}

\begin{algorithm}[t]
\small
\caption{S-HCI-GIBO}\label{algo:s-hci-gibo}
\begin{algorithmic}[1]
    \State \textbf{Hyperparameters}: hyperpriors for GP hyperparameters,
    improvement confidence $\alpha$,
    SimToReal threshold $\beta$,
    stepsize $\eta$,
    \State $\hat\theta_* \leftarrow \theta_0$, $\mathcal{D}  \leftarrow \mathcal{D}_0$
    \Repeat \Comment{Policy updates}
    \State $\IS \leftarrow \IS_{\simu}$
    \Comment{Start with $\IS_{\simu}$}
    \Repeat
    \Comment{DoE for descent direction}
    \State $\theta_i \leftarrow \textsc{Query}(\mathcal{D}, \hat{\theta}_*, \IS)$
    \If{$\IS$ is $\IS_{\simu}$}
    \State $y_i \leftarrow f_{\simu}(\theta_i)+\epsilon_i$
    \Else
    \State $y_i \leftarrow f(\theta_i)+\epsilon_i$
    \EndIf
    \State $\mathcal{D}  \leftarrow \mathcal{D} \cup (\theta_i, y_i, \IS)$
    \If{$\textsc{SimToReal}(\mathcal{D}, \hat\theta_*, \IS) \leq \beta$ }
    \State $\IS \leftarrow \IS_{\real}$ \Comment{Switch from simulator to robot}
    \EndIf
    \Until{$\textsc{Commitment}(\mathcal{D}, \hat\theta_*, \IS_{\real}) \geq \alpha$}
    \State $\mu^\prime_* \leftarrow \textsc{DescentDirection}(\mathcal{D}, \hat\theta_*, \IS_{\real})$
    \State $\hat\theta_* \leftarrow \hat\theta_* - \eta \, \mu^\prime_*$ \Comment{Optional: normalized gradient descent}
    \Until{$\textsc{Termination}(\mathcal{D}, \hat\theta_*)$}
\end{algorithmic}
\end{algorithm}

Simulators are ubiquitous, fast, and play a vital role in exposing robots to learning tasks in the safety of the virtual world. In the following, we first give a high-level overview of S-HCI-GIBO, and then introduce the algorithm's details.
\subsubsection{S-HCI-GIBO Algorithm}
S-HCI-GIBO, as summarized in Alg.~\ref{algo:s-hci-gibo}, follows the general GIBO framework with its two nested loops but includes two information sources that can be queried in the inner loop.
In this algorithm, we use a binary indicator \(\IS\) and denote the simulator and the real robot by \(\IS_{\simu}\coloneqq 0\) and  \(\IS_{\real} \coloneqq 1\), respectively.
S-HCI-GIBO will \textsc{Query} the simulator first until the \textsc{SimToReal} function reaches a preset threshold.
The threshold indicates data from the simulator is no longer significant for the \textsc{Commitment} condition.
Then, the optimizer will \textsc{Query} the robot.
When the threshold in \textsc{Commitment} is reached, the optimizer leaves the inner loop. Note that it is possible to achieve the \textsc{Commitment} condition only with data from a simulator.

We propose a dual information source derivative GP model to model \(f\) and \(f_{\simu}\) and extend the GI for a dual information source as the \textsc{Query} function. 
The \textsc{Commitment} and \textsc{DescentDirection} functions are identical to HCI-GIBO. The dual information source derivative GP model and the \textsc{SimToReal} switching rule are introduced in the following.

\subsubsection{Dual information source derivative GP model}
Following the model \eqref{eq:model_GIBO}, and motivated by Poloczek \textit{et al.} \cite{poloczek2017multi}, we construct a model with a distinctive kernel that enables the transition of queries between different information sources.
We extend the parameter vector \( \theta\) by a binary indicator \(\IS\). Subsequently, we introduce the tuple \(x = ( \theta, {\IS})\).
%, and the joint distribution at \(\hat \theta_*\) is  
% \begin{equation}
% \begin{aligned}
% 	&\begin{bmatrix} y \\ \nabla F(x_*)
% 	\end{bmatrix} \sim \\
% 	&\mathcal{N} \left( \begin{bmatrix}
% 		\mu \left(\queriesext \right) \\ \nabla \mu (x_*)
% 	\end{bmatrix},
% 	\begin{bmatrix}
% 		K\left(\queriesext, \queriesext \right) + \sigma^2 I  &  \nabla K\left(\queriesext, x_*\right) \\
% 		 \nabla K\left(x_*, \queriesext\right) &  \nabla^2 K( x_*, x_*) \\
% 	\end{bmatrix}
% 	\right),
% \end{aligned} \label{eq:model}
% \end{equation}
% where \(y\) includes \(n\) observations, and \(\queriesext = \left[x_0, x_1, \cdots, x_n \right]\) is a vector of location pairs of these observations.
The covariance matrix \(K\left(\queriesext, x_*\right)\) has the form \cite{marco2017virtual}
\begin{equation}
K\left(\queriesext, x_*\right) = K_f(\theta,  \hat\theta_*) + k_\delta({\IS}, {\IS}_*) K_{m}( \theta,  \hat\theta_*). \label{eq:new_kernel}
\end{equation}
In \eqref{eq:new_kernel}, \(K_f\) and \(K_m\) are given by two kernel functions $k: \Theta \times \Theta \mapsto \mathbb{R}$, in which the kernel hyperparameters can be set respectively. \(k_\delta({\IS}, {\IS}_*) = {\IS}\cdot{\IS}_* \), and is equal to one if both parameters indicate \(\IS_{\real}\) and zero otherwise. Given the derivative GP model \eqref{eq:model_GIBO}, conditioning the GP's prior distribution on \(\{\queriesext, \bar{y}\}\) leads to the posterior distribution, which can be readily obtained by substituting \(K(\queries,  \hat \theta_*)\) into \eqref{eq:posterior_derivative_gp} with the new kernel \(K(\queriesext, x_*)\) in \eqref{eq:new_kernel}.

\subsubsection{Sim-to-real switching rule}
We implement the GI acquisition function for both information sources.
For any parameter $\theta$, there should now exist points on both \(\IS_{\simu}\) and \(\IS_{\real}\) that are informative for its respective gradients.
Thus, at each timepoint the algorithm can decide to query either the simulator or the real robot.
Experiments on the real robot are typically more informative than simulations.
Thus, when maximizing gradient information without consideration of the cost, the algorithm will mostly neglect the simulator.
However, the information from the simulator is cheap as it requires no interaction time with the robot.
We therefore establish a hierarchy and first exhaust gradient information from the simulator before turning to the real robot.

Given a derivative GP model, the data-set \(\mathcal{D}\) of the observation, and the current guess of the best policy \( \hat \theta_*\), we revisit and rewrite the utility in \eqref{eq:gi_acquistion} as the expected difference between the derivative's variance \emph{before} and \emph{after} observing a new point \((x, y)\), which contains the policy \(\theta\), the information source \(\IS\), and the performance observation \(y\)
\begin{equation} \label{eq:acq}
\begin{aligned}
&\alpha_{\rm{S}-\rm{GI}}\left(x \mid \mathcal D, (\hat\theta_*, {\IS}_{\real})\right) \\
&= \E \bigl[\Tr \bigl(\Sigma_*^\prime \mid \mathcal{D}\bigr)
- \Tr\bigl(\Sigma_*^\prime \mid \mathcal{D}\cup\left(x, y\right) \bigr)\bigr],
\end{aligned}
\end{equation}
where \(\Sigma_*^\prime\) is the variance of the derivative GP model at the pair \((\hat\theta_*, \IS_{\real})\).
Note that we aim to reduce the gradient uncertainty of the policy on the real robot, and therefore $\alpha_{\rm{S}-\rm{GI}}$ depends on the GP variance for ${\rm IS}_{\rm real}$. 
The posterior at the policy \(\hat\theta_*\) can be conditioned on data from both \(f\) and \(f_{\simu}\).

% The covariance function is independent of the observed targets \(y\) \cite{muller2021local}, and the optimization over the expectation can be written as 
% \begin{equation}
% \begin{aligned}
% &\argmax_\theta \alpha_{\rm{S}-\rm{GI}} = \argmin_{\theta} \Tr\left(\Sigma^\prime_*\mid \queriesext\cup x\right).
% \end{aligned} \label{eq:acq}
% \end{equation}
% %where we define a virtual data-set \(\hat { X} = \left[( \theta_0, \IS_0), \cdots,  \right.\) \(\left. ( \theta_n, \IS_n),  ( \theta, \IS)\right]\coloneqq [ X, ( \theta, \IS)]\).
% Since there are two information sources, \ie \(\IS = \IS_{\rm real}\ {\rm or}\ \IS_{\rm sim}\), each source has a most informative new parameter \( \theta\) to query by switching \(\IS\) in (\ref{eq:acq}), and the new parameter is only dependent on where we query next. With the condition on \(\IS\), we define
% \begin{equation} \label{eq:point_on_sim}
%  \textsc{Query}(\mathcal{D},\hat\theta_*, {\rm IS}) \coloneqq \argmax_{ \theta} \alpha_{\rm{S}-\rm{GI}}.
% \end{equation}
% Similar to \eqref{eq:gi_acquistion}, we have
% \begin{equation}
% \begin{aligned} \label{eq:utility}
% &\alpha_{\rm{S}-\rm{GI}} = \Tr \left( \nabla K\left(x, \hat\queriesext\right) \right.\\
% &\left.\left(K\left(\hat\queriesext, \hat\queriesext\right)+\sigma^2 I\right)^{-1} \nabla K\left(\hat\queriesext, x\right) \right),\\
% &\hat{\queriesext} = [\queriesext, x], \ \ {\IS} = {\IS}_{\rm real}\ {\rm or}\ {\IS}_{\rm sim}.
% \end{aligned}
% \end{equation}
Compared to acquisition function GI \eqref{eq:gi_acquistion}, \(\alpha_{\rm{S}-\rm{GI}}\) calculates the gradient information for $\IS_{\real}$ from a policy executed on either a real robot or a simulator. Note that for the same policy \(\hat \theta_*\), \(\alpha_{\rm{S}-\rm{GI}}\) varies in terms of the selected information source.

We are now ready to introduce a threshold to indicate when the algorithm should query the real robot instead of the simulator. 
This can be quantified by measuring the utility function for the neighboring query steps on the simulator.
Consider queries in one gradient step.
We initially execute our queries on the simulator.
This approach enables us to fully exploit the gradient information gained from the simulator, and effectively decreases the need for samples from the real robot.
Given the query step \(i\), the data set $\mathcal{D}$, and the current guess of the best policy \(\hat \theta_*\). Then the difference between the utility at step \(i\) and \(i+1\) on the simulator is
\begin{equation}
\begin{aligned} \label{eq:simtoreal_switching}
\textsc{SimToReal} \coloneqq & \alpha_{\textrm{S-GI}}\left(\left(\theta_{i+1}, \textrm{IS}_{\textrm{sim}}\right)\mid \mathcal D, (\hat\theta_*, {\IS}_{\real})\right) \\
- &\alpha_{\textrm{S-GI}}\left(\left(\theta_{i}, \textrm{IS}_{\textrm{sim}}\right)\mid \mathcal D, (\hat\theta_*, {\IS}_{\real})\right)
\end{aligned}
\end{equation}
This measures how much the variance will change for the gradient if such query point is included in the dataset. We introduce a threshold \(\beta\) as a hyperparameter for the sim to real switching. $\textsc{SimToReal} \leq \beta$ suggests that data from \(\IS_{\simu}\) reduces the uncertainties for the \textsc{DescentDirection} with a variance change of no more than the threshold $\beta$, so that the data from the simulator is no longer significant to achieve high confidence improvement for the policy on the real robot.
%While $\beta$ is a hyperparameter, the utility calculated by \eqref{eq:acq} can provide a rough estimate of how small $\beta$ should be, for example, $\beta$ could be 5\% of the utility.
When $\beta$ is small, more queries on the simulation will be included, and it could possibly further decrease the number of samples needed from the real robot.
However, this will increase overall training time.
%One could maintain a balance between sample reduction and training time by properly setting the threshold.
%We define \eqref{eq:simtoreal_switching} as the SimToReal switching rule.

\section{Toy examples for intuition and visualization}
\begin{figure}
    \centering
    \includegraphics[width=3.5in]{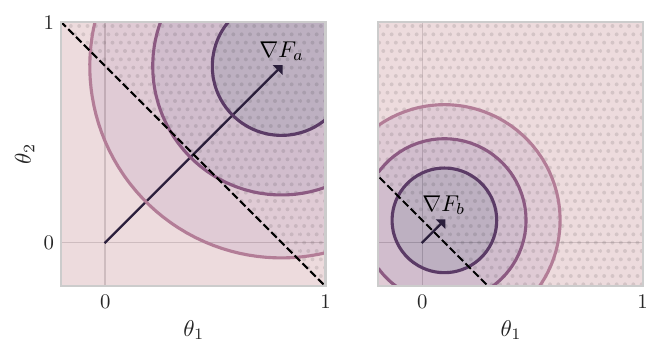}
    \caption{\textbf{Visualization of improvement confidence regions:} Improvement confidence of gradient distribution \(\nabla F_a\) and \(\nabla F_b\) are \SI{97}{\percent} and \SI{76}{\percent}, respectively. Mean vectors are denoted by black arrows, and the contour lines show the density of the gradient distribution. If \(L\eta = 1\) and true gradients are in regions filled with dots, gradient step using the mean vector improves the policy.}
    \label{fig:confidence}
\end{figure}
\begin{figure}
\centering
\begin{tikzpicture}
\node[inner sep=0pt] (step0) at (1.5, 0)
    {\includegraphics[width=.27\textwidth]{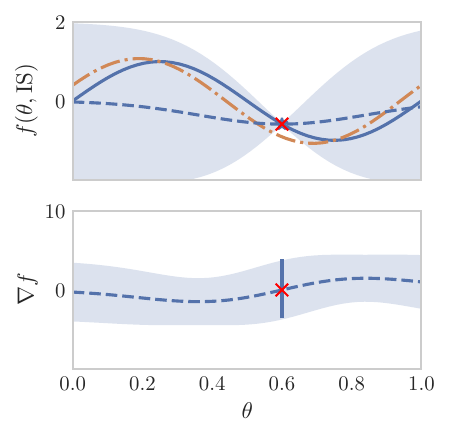}};
\node[inner sep=0pt] (sim) at (-0.6, -5.)
    {\includegraphics[width=.26\textwidth]{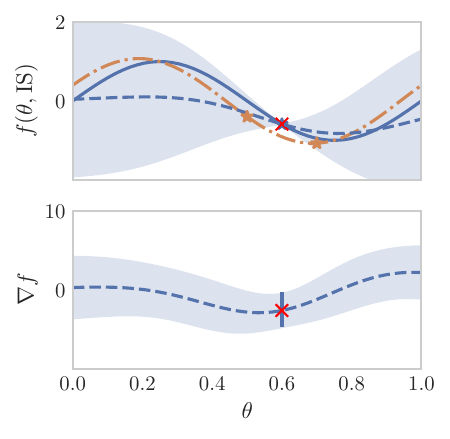}};
\node[inner sep=0pt] (real) at (3.8, -5.04)
    {\includegraphics[width=.235\textwidth]{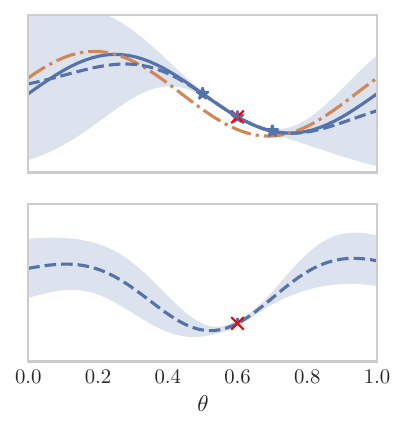}};
\draw[->,thick] (0.8, -2) -- (-0.4, -2.8)
    node[midway,fill=white] {\small \textcolor{sim}{query from $f_{\simu}$}};
\draw[->,thick] (2.6, -2) -- (3.8, -2.8) %ugly hard coding but it works
    node[midway,fill=white] {\small \textcolor{real}{query from $f$}};
\end{tikzpicture}
    \caption{\textbf{Multi-fidelity queries:} Comparison of uncertainties (blue shaded regions) after queries at same locations ($0.5$ and $0.7$ shown with star markers) from the simulator and the robot. Use toy objective function: \textcolor{real}{$f$: blue line} and \textcolor{sim}{$f_{\simu}$: orange dash-dotted line}; \textcolor{real}{Posterior mean $\mu$ and $\nabla \mu$): blue dashed line}; Current guess of the best policy: red cross marker.}
    \label{fig:mis}
\end{figure}
In this section, we present visualization examples to help develop intuition into how the HCI-GIBO and S-HCI-GIBO algorithms work. To this end, we first illustrate the improvement confidence for an abstract gradient example and then show how an additional biased information source can reduce variances in estimates of the objective function and its gradient. Equipped with the above properties, we demonstrate the optimization process for a simple one-dimensional objective.

\subsection{Intuition into \theo\ref{thm:improvement_confidence}}

\theo\ref{thm:improvement_confidence} defines a probability of improvement that only depends on the belief about the gradient. We can therefore generate some toy gradient examples without setting an explicit performance function. Here, we select
\begin{equation}
\label{eq:prob_dist}
\begin{aligned}
    &\nabla F_a  \sim \mathcal{N} \left(
    \left[
    \begin{array}{*{2}{@{}C{\colwd}@{}}}
        0.8 \\ 0.8        
    \end{array}
    \right],
    \begin{bmatrix*}[r]
        0.3 & 0 \\
        0 & 0.3
    \end{bmatrix*}
    \right), \\
    &\nabla F_b  \sim \mathcal{N}
    \left(
    \left[
    \begin{array}{*{2}{@{}C{\colwd}@{}}}
        0.1 \\ 0.1        
    \end{array}
    \right],
    \begin{bmatrix*}[r]
        0.1 & 0 \\
        0 & 0.1
    \end{bmatrix*}
    \right).
    \end{aligned}
\end{equation}
In this example, we assume \(L\eta = 1\). This allows us to easily interpret \eqref{eq:criteria}. The left side of \eqref{eq:criteria} gives the signed magnitude of a projection (true gradient \(\nabla f\) on the descent direction \(\direction\)). With the assumption \(L\eta = 1\), we know that this magnitude should be larger than the half magnitude of the descent direction \(\direction\).

In \fig\ref{fig:confidence}, we show the regions (dots) in which the projections of unknown true gradients on its mean vector (arrows) fulfill \eqref{eq:criteria}. Note that the probability density of true gradients is given in \eqref{eq:prob_dist} and marked with contour lines in the figure. Improvement confidence is calculated analytically by \eqref{eq:cdf} and \eqref{eq:dist}. \(\nabla F_a\) has higher improvement confidence than \(\nabla F_b\) although it has higher uncertainties. If we set a probability threshold \(p\), \eg \(p = 0.9\), the optimizer would further query information sources to improve the confidence for the case \(\nabla F_b\). However, for the case \(\nabla F_a\), it would stop querying and use the gradient to update the policy with sufficiently high improvement confidence. Comparing these two cases reveals that it is possible to reduce queries when the objective function is ``steep,'' even with higher uncertainties. At the same time the optimizer has to be more cautious, \ie incorporate more samples, to reduce uncertainties for small gradients.
\begin{figure*}[t]
    \centering
    \begin{subfigure}{0.22\textwidth}
         \centering
         \includegraphics[width=\linewidth]{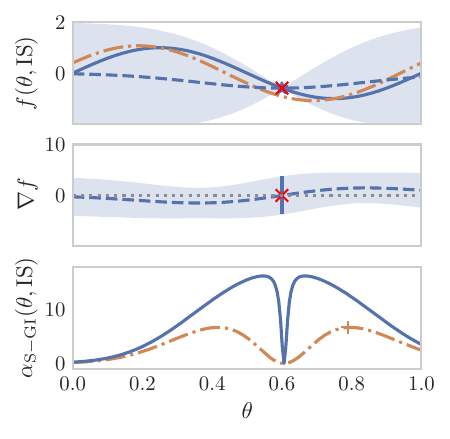}
         \caption{\label{fig:demo0} initial policy}
    \end{subfigure}
    \hspace*{-1em}
    \begin{subfigure}{0.205\textwidth}
         \centering
         \includegraphics[width=\linewidth]{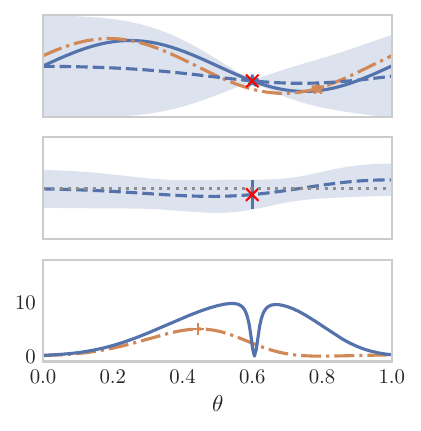}
         \caption{\label{fig:demo1} query from $f_{\simu}$}
    \end{subfigure}
    \hspace*{-1em}
    \begin{subfigure}{0.2\textwidth}
         \centering
         \includegraphics[width=\linewidth]{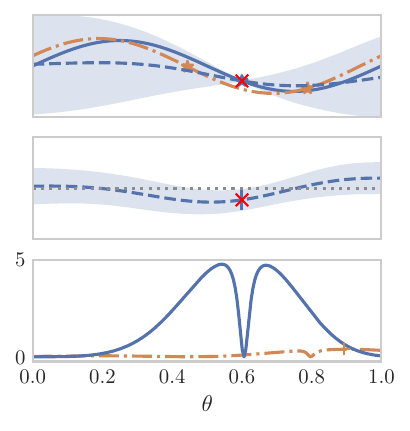}
         \caption{\label{fig:demo2} query from $f_{\simu}$}
    \end{subfigure}
    \hspace*{-1em}
    \begin{subfigure}{0.2\textwidth}
         \centering
         \includegraphics[width=\linewidth]{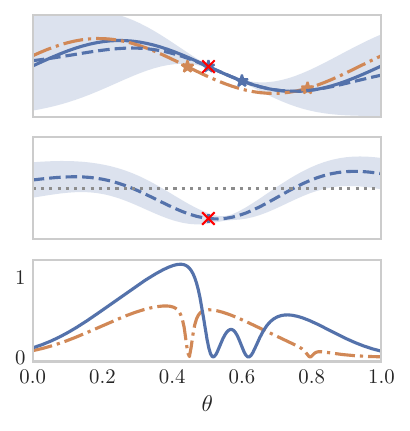}
        \caption{\label{fig:demo3} update policy}
    \end{subfigure}
        \hspace*{-1em} 
    \begin{subfigure}{0.2\textwidth}
         \centering
         \includegraphics[width=\linewidth]{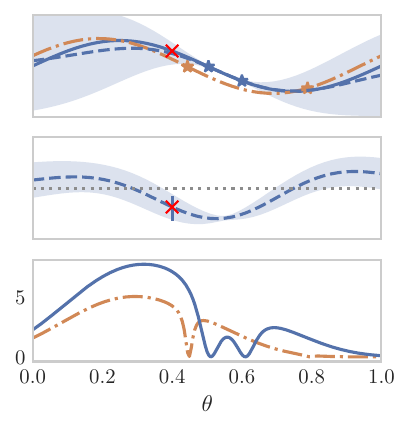}
         \caption{\label{fig:demo4} update policy}
    \end{subfigure}
    \caption{\textbf{Visualization of the S-HCI-GIBO optimization process with a 1-dimensional function.} Top: The objective functions of \textcolor{real}{\(f\) (blue line)} and \textcolor{sim}{\(f_{\simu}\) (orange dash-dotted line)}. The red cross symbol refers to the current parameter \(\theta_i\). The blue and orange star markers represent the samples that have been queried from \(f\) and \(f_{\simu}\). The posterior mean \textcolor{real}{\(\mu( \theta, \IS_{\real})\)} is shown with the blue dashed line. The shaded regions show the standard deviation. Middle: The posterior mean conditioned on star markers. Bottom: The acquisition function, \ie the gradient information of the \textcolor{real}{\(f\) (blue line)} and the \textcolor{sim}{\(f_{\simu}\) (orange dash-dotted line)}.}
    \label{fig:demo}
\end{figure*}
\subsection{Example update from a biased information source}
To demonstrate how the dual information source derivative GP model accounts for the bias in samples from \(f_{\simu}\), we consider an example where the objective function is \(f = \sin(2\pi\theta)\), and we have an extra information source \(f_{\simu} = f + 0.4\cos(2\pi\theta)\).
We plot both $f$ and $f_{\simu}$ represented by blue and orange dash-dotted lines, respectively, in Fig. \ref{fig:mis} (top).
Additionally, the graph also includes the posterior of $f$ and $\nabla f$.
We assume Gaussian noise here with a mean of 0 and a variance of 0.01.
We take as initial parameter \(\theta = 0.6\). 
In the following, we consider the squared exponential kernel for \(K\) in \eqref{eq:model_GIBO} and \(K_f, K_m\) in \eqref{eq:new_kernel}.% any twice differentiable kernel would be compatible with the framework.

We subsequently explore the posterior of the GP and its derivatives by adding two testing queries (\(\theta = 0.5\ {\rm and}\ 0.7\)) to \(f_{\simu}\) (\fig\ref{fig:mis}, bottom left) and \(f\) (\fig\ref{fig:mis}, bottom right). 
The blue shaded regions, which represent the uncertainty of $\nabla f$, shrink in the bottom part of the figure compared to the top.
This demonstrates the data from $f$ and the biased data from $f_{\simu}$ can both reduce the uncertainty of the gradient estimates.
Further, if we compare the bottom right figure with the bottom left figure, the data from $f$ contributes more to this reduction compared to that from $f_{\simu}$.
Such comparison motivates us to think carefully about the trade-off between cheap data and acquired information.
%We can update the model even though the sample is from a biased information source. 
%However, such a sample from \(f_{\simu}\) has more uncertainty than one from the actual objective \(f\). This is also reflected in the uncertainty on the gradients.

\subsection{Visualization of Algorithm~\ref{algo:s-hci-gibo}}
\fig\ref{fig:demo} shows a simple one-dimensional objective function and illustrates how the information from two different information sources (highlighted in blue and orange) changes the posterior of the gradients. 
Here, we proceed with the same initial conditions presented in the top panel of \fig\ref{fig:mis}, where we use the same objectives: a sine objective and a biased one. 
The optimal policy parameter is $\theta_* = 0.25$.
As can be seen from the acquisition function in \fig\ref{fig:demo0}, queries from \(f\) can reduce gradient uncertainty more than queries from \(f_{\simu}\). 
In addition, since the simulator leaves more uncertainty after evaluation, the optimum of the \(f_{\simu}\) acquisition function is farther from $\hat{\theta}_*$ than that of \(f\). 
Intuitively, to reduce uncertainty of gradient estimates, queries must be made farther away to eliminate the bias of the data.
Starting with $\hat{\theta}_* = 0.6$, the optimizer queries the simulator twice, shown in \fig\ref{fig:demo1} and \fig\ref{fig:demo2}. 
After each query, there is a reduction in the variance at $\hat{\theta}_*$, and in \fig\ref{fig:demo2} we assume the high confidence improvement condition has been reached.
Then S-HCI-GIBO updates the policy twice in \fig\ref{fig:demo3} and \fig\ref{fig:demo4}.
Further, from \fig\ref{fig:demo3} to \fig\ref{fig:demo4}, S-HCI-GIBO can update the policy without querying any information sources if there is high confidence in policy improvement.

\section{Synthetic experiments}
We next demonstrate the data-efficiency of HCI-GIBO and S-HCI-GIBO in large-scale synthetic experiments\footnote{The code of HCI-GIBO and S-HCI-GIBO is based on the code by M{\"u}ller \etal~\cite{muller2021local} and available at  \href{https://github.com/Data-Science-in-Mechanical-Engineering/hci-gibo}{https://github.com/Data-Science-in-Mechanical-Engineering/hci-gibo}.}.
In particular, we compare both algorithms to: BO with expected improvement (EI-BO)~\cite{jones1998efficient},
augmented random search (ARS)~\cite{mania2018simple}, gradient information Bayesian optimization (GIBO), and confidence region Bayesian optimization (CRBO) \cite{frohlich2021cautious}; and demonstrate that they are significantly faster in finding good solutions, including in high-dimensional domains.
Further, our synthetic experiments double as the ablation study: 
comparisons among ARS, EI-BO, GIBO, HCI-GIBO, and S-HCI-GIBO help us evaluate the effectiveness of gradient approaches, acquisition functions, improvement confidence, and data from simulator, respectively.

\begin{figure*}[!t]
    \centering
    \includegraphics[width=7.2in]{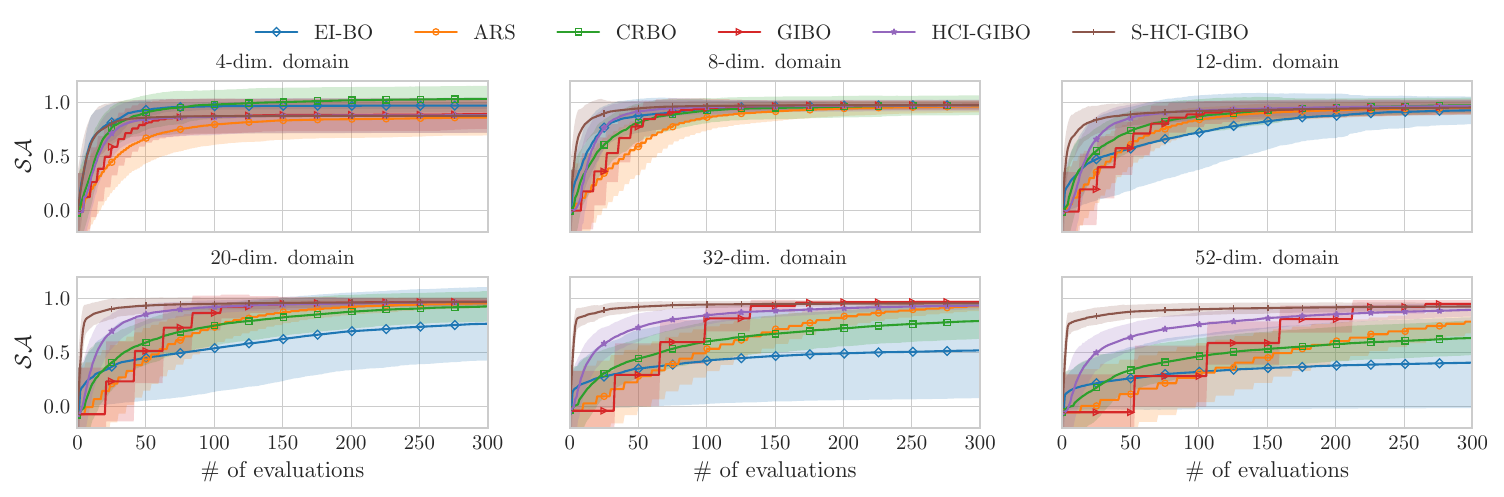}
    \caption{\textbf{Within-model comparison:} Mean of the reward for six different dimensional function domains for 100 runs. Evaluations are for $f$ and do not include queries to $f_{\mathrm{sim}}$. HCI-GIBO and S-HCI-GIBO show greater data efficiency, especially in a high dimensional setting.}
    \label{fig:within-model}
\end{figure*}
\subsection{Experiment Setup}
Rather than testing performance on hand-crafted objective functions, we sample the objective from a known GP prior. This approach is called within-model comparison \cite{hennig2012entropy} and is commonly used as a benchmark for BO. In this way, the objective functions not only satisfy Assumption \ref{ass:lipschitz}, but can also be generated on various dimensions to evaluate the scalability of the methods. 
Such a setting helps to demonstrate that the method works under ideal conditions, and when these experiments are combined with the following robot learning experiments, it helps us to analyze the discrepancy between its ideal and practical performance, and therefore helps us to evaluate how restrictive Assumption 1 is.
We analyze synthetic benchmarks for up to 52 dimensions.

The experiments are carried out over a \(d-\)dimensional unit domain \(I=[0, 1]^d\). We generate 100 different test functions for each domain, and for each function we jointly sample 1000 values from a GP prior with an SE kernel and unit signal variance using a quasi-random Sobol sampler. Then, we take the mean of the GP posterior as an objective function so that hyperparameters are known.
%For the S-HCI-GIBO, a real performance function \(f\) is sampled from a GP model and the function \(f_{\simu}\) on the simulator is generated by adding \(f_{\rm gap}\) (also sampled from a GP model) to the performance \(f\).

We increase lengthscales of GPs in higher dimensions so that those high-dimensional objective functions vary less over a fixed distance, and are therefore easier to model and predict.
The difficulties of the optimization problem are thus comparable over all domains.
The lengthscale distribution for different dimensional domains is available in Appendix A.5 of~\cite{muller2021local}.
For the S-HCI-GIBO, \(f\) and \(f_{\rm gap}\) are sampled independently from a GP model using the same lengthscale.
However, the outputscale of the GP model for $f_{\rm gap}$ is 20\% of that for $f$. 
All algorithms are started in the middle of the domain \(x_0 = [0.5]^d\) and have a limited budget of 200 noisy function evaluations. The noise is normally distributed with standard deviation \(\sigma = 0.1\). 

We compare the proposed algorithms HCI-GIBO and S-HCI-GIBO to GIBO, CRBO, ARS, and EI-BO with expected improvement as acquisition function, and we use the SE kernel for all GP-based methods. Domain knowledge is passed to the ARS algorithm by scaling the space-dependent hyperparameters with the mean of the lengthscale distribution. For CRBO, there is a hyperparameter \(\gamma \in (0, 1]\) that governs the effective size of confidence regions. In our experiment, \(\gamma\) is selected as $0.6$ to achieve the best average performance.
\subsection{Results}
For the synthetic experiment, we rescale the reward in the range of $[0, 1]$ and define a solution accuracy metric \(\mathcal{SA} = {\obj(\hat \theta_*)}/{\obj(\theta_*)}\).
Here, \(\obj(\theta_*)\) is the optimal performance found by an exhaustive brute-force approach. 
Therefore the optimal solution is at $\mathcal{SA} = 1$.
Fig. {\ref{fig:within-model}} shows the solution accuracy (y-axis) that can be reached after a fixed number of queries on the real function (x-axis).
Since we assume samples from the simulator are ``free'' the x-axis for S-HCI-GIBO only shows the number of queries from $\obj$ but not the additional queries to $\obj_\mathrm{sim}$.
The standard deviation is included in the plot over 100 objective functions per domain. The within-model comparison shows that overall HCI-GIBO and S-HCI-GIBO exhibit better data efficiency in the high dimensional setting.

Comparing GIBO with ARS, we find that the acquisition function GI \eqref{eq:gi_acquistion} improves data efficiency by maximizing the gradient information contained in each query. HCI-GIBO and S-HCI-GIBO also benefit from this ability since they also use it for active learning. 

If we now turn to the comparison between HCI-GIBO and GIBO, we observe that HCI-GIBO uses less queries (interaction time) to reach the same solution accuracy as GIBO, and this phenomenon is much more pronounced in high dimensional search spaces. This finding is significant in confirming \emph{(i)} HCI-GIBO can query on demand to establish confidence in policy improvement, instead of querying a fixed number of samples; \emph{(ii)} HCI-GIBO is suitable for efficiently finding solutions of high dimensional tasks, which are prevalent in the robotics community. Further, compared with the setting without \(\IS_{\simu}\), S-HCI-GIBO was able to leverage the information from the biased simulator and find better solutions with less data compared to the other methods. 

\begin{figure*}[!t]
    \centering
    \includegraphics[width=7.2in]{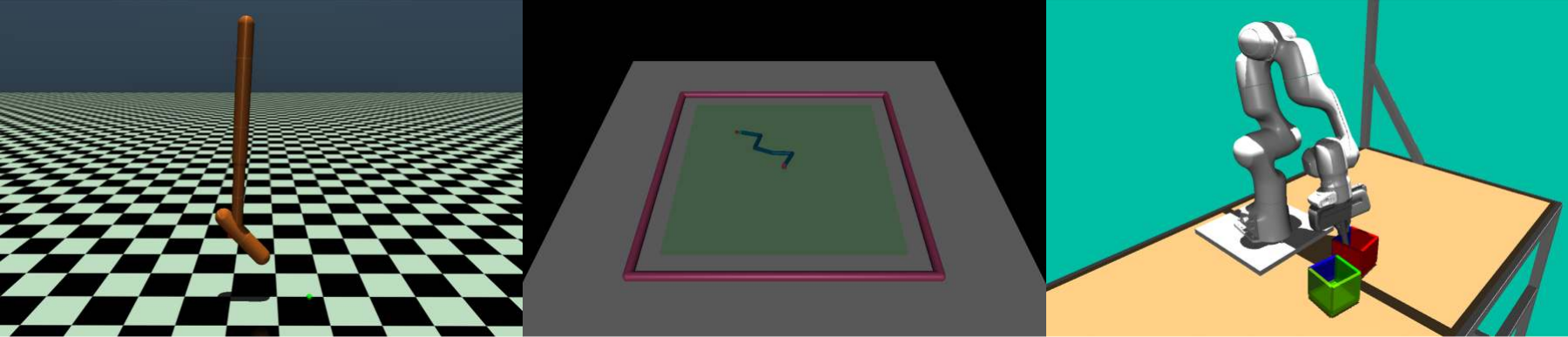}
    \caption{\textbf{Visualization of three \emph{fancy\_gym} environments:} hopper jumping (15-dim. domain), reacher (25-dim. domain), and box pushing (70-dim. domain).}
    \label{fig:gym_env}
\end{figure*}
\begin{figure*}[!t]
    \centering
    \includegraphics[width=7.0in]{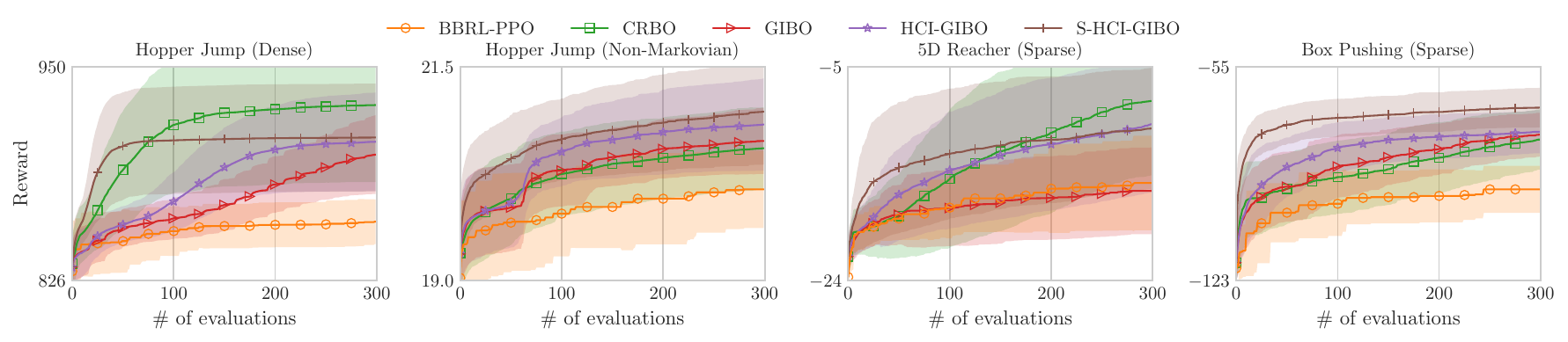}
    \caption{\textbf{Results of fine-tuning control policies for three \emph{fancy\_gym} environments:} Evaluations start from five random initial conditions over 20 runs. BO approaches exhibit superior refinement in task performance compared to the black-box RL.}
    \label{fig:gym_results}
\end{figure*}

\section{Robot learning experiments}\label{sec:robot}
We evaluate our proposed approaches (HCI-GIBO and S-HCI-GIBO) in three simulated \emph{fancy\_gym}~\cite{fancy_gym} scenarios. Additionally, we have assessed the performance of the proposed approaches on robotic hardware\footnote{A video is available at \href{https://www.youtube.com/watch?v=9iSUzwjN7ko}{https://www.youtube.com/watch?v=9iSUzwjN7ko}.} by applying them to the setup shown in \fig\ref{fig:sim-and-real}.
As discussed in Section~\ref{sec:intro}, we use MPs as policy function approximators.

In the \emph{fancy\_gym} experiments, we consider fine-tuning of MPs generated by a pre-trained deep RL policy using the algorithm proposed by \citet{otto2023deep}.
In this setting, MP-generated signals are joint-space trajectories, which are transformed into sequential actions using a trajectory tracking controller, and the specific controller chosen and implementation are from the \emph{fancy\_gym} package.
We assume a scenario where the NN policy has been pre-trained using the simulator but fails to meet satisfactory performance requirements on the real system. 
Through data-efficient fine-tuning that still incorporates the simulator, our proposed approaches enable task performance tailored to repeated tasks. 
In these experiments, we aim to demonstrate that BO approaches show data efficiency in fine-tuning the RL agent.

\begin{figure*}[!t]
    \centering
    \includegraphics[width=\textwidth]{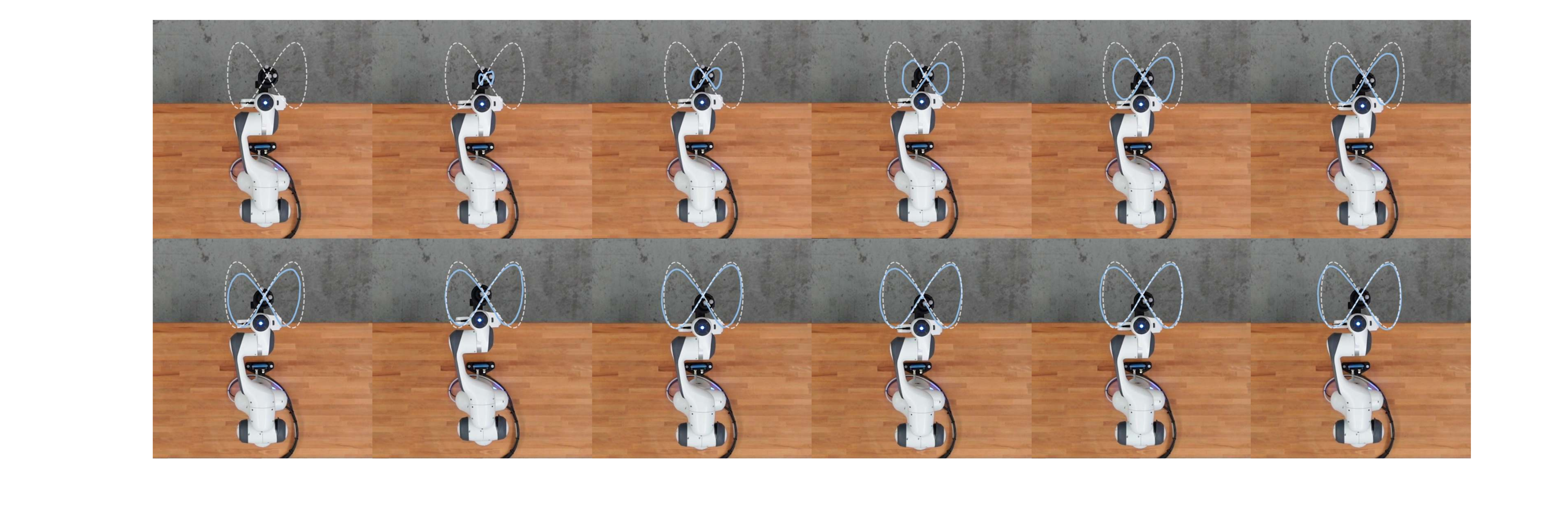}
    \caption{\textbf{Experimental results:} The trajectory is improved after each policy update. The reference and the actual trajectory are marked by white-dotted line and blue line, respectively.}
    \label{fig:improved_trajectory}
\end{figure*}

In our hardware experiment, we consider an impromptu trajectory tracking problem: an inverted pendulum is balanced by a robotic arm, and the tip of the pole is supposed to follow the impromptu trajectory (see \fig\ref{fig:improved_trajectory}). The MP-generated signal can be regarded as a persistent external disturbance to the closed-loop pendulum control system, prompting the tip of the pendulum to leave its equilibrium point. 
We use the proposed methods to tune the MP-generated signal so that high trajectory tracking precision can be achieved in the presence of unknown dynamics.
In this experiment, we show the proposed methods are capable of efficiently learning and improving an impromptu trajectory from scratch, in particular with unknown dynamics, for a robotic system.

\subsection{Fine-tuning Deep RL Agents}
The following experiments showcase the potential of our method to fine-tune MP-based policies data-efficently. Generally, the MP can be the outcome of expert design or imitation learning. In practice, fine-tuning of MPs can help a system adjust to new tasks, systems or adapt in time-varying environments. Here, we use the method of \citet{otto2023deep} to generate a candidate MP and fine-tune its parameters to a fixed task. The results show that local BO is able to improve the candidate solution with relatively few evaluations, especially when compared to BBRL-PPO. Incorporating an additional information source generally further improves data-efficiency.

\paragraph{Environment setup}
We consider the \emph{Hopper Jump}, \emph{5D Reacher}, and \emph{Panda Box Pushing} environments from the \emph{fancy\_gym} (see \fig\ref{fig:gym_env}).
%The action spaces of Hopper Jump (use dense reward setting) default to 15 and are extended to 60 dimensions (use non-Markovian reward setting).
%For the 5D Reacher and the Box Pushing task, the action spaces are 25 and 70 dimensions, respectively, and the reward implementation in these two is sparse.
We first train agents with the Black-Box RL Proximal Policy Optimization (BBRL-PPO) \cite{otto2023deep}.
The RL policies are multi-layer NNs that map a context to weights of MPs which in turn generate a trajectory.
For the fine-tuning experiment, we fix the context, \eg goal position and initial states of the agent, and take the MP generated by the pre-trained NN as the starting point and optimize the weights of the probabilistic MP (ProMP) \cite{paraschos2013probabilistic}, \ie the action (trajectory) of the agent.
To evaluate the effect of an additonal information source we design an artificial ``real'' reward signal by adding a sim-to-real offset sampled from a known GP prior following the procedure in our synthetic experiments.
With this setting, we mimic the situation where the robot initially learns a general policy in simulation through RL, then the robot's performance is fine-tuned to ensure it meets performance requirements on the actual tasks. 

We compare the performance of BO approaches: HCI-GIBO, S-HCI-GIBO, GIBO and CRBO to a black box RL method, namely BBRL-PPO \cite{otto2023deep}.
The unknown hyperparameters of BO approaches are hand-tuned.
Specifically, we normalize the action and the reward, and set lengthscale and outputscale to 0.1 and 1, respectively.
The improvement confidence $\alpha$ and the stepsize $\eta$ are $0.9$ and $0.2$, respectively.
The SimToReal threshold $\beta$ is set to 5 for the S-HCI-GIBO.
For CRBO, the effective size of confidence regions $\gamma$ is set to 0.2 for the Hopper Jump, and 0.1 for the other two environments to cater for the optimal overall performance.
For the BBRL-PPO, we took the hyperparameters from the appendix in \cite{otto2023deep}.

For each \emph{fancy\_gym} experiment, we randomly choose an initial context, and took the MP generated by the BBRL-PPO for the initialization point of the optimization.
We then run the optimization for each approach twenty times using the artificial ``real'' reward.
During the optimization, S-HCI-GIBO is still able to access the simulation.
We compare the BO approaces to BBRL-PPO by continuing the training from the existing checkpoint using the fixed context.
Each method has a budget of 300 queries to the artificial ``real'' reward.

\paragraph{Results}
The rewards achieved by these approaches are reported in \fig\ref{fig:gym_results}.
We find that both HCI-GIBO and S-HCI-GIBO
show an enhancement over the GIBO baseline performance.
While CRBO eventually achieves higher reward than GIBO and its variants when dealing with 15- and 25-dimensional problems, we notice a clear trend indicating that CRBO's scalability relative to the dimensionality of the policy space is inferior to that of GIBO and proposed methods: it does not compete with proposed methods (HCI-GIBO and S-HCI-GIBO) for the 60 and 70-dimensional policy domain.
Meanwhile GIBO and its variants have a smaller variance across multiple optimizations compared to CRBO.
Further, it is remarkable that for all fine-tuning tasks, S-HCI-GIBO consistently outperforms the others in terms of early improvements, making it the quickest to adapt and showcasing its potential for fast task adaptation in robot learning.

On the other hand, we also observe that the performance of GIBO and its variants is not quite as dominant, when contrasted with results from the synthetic experiments and compared to CRBO.
We attribute this performance decline to inaccuracies in hyperparameters.
In these experiments, objectives to be optimized are likely to violate Assumption \ref{ass:lipschitz}, and hyperparameters such as lengthscale of the GPs can only be estimated and hand-tuned.
These insights show the role that accurate hyperparameters settings play in the successful deployment of our proposed methods. Despite these problems and CRBOs robust performance the proposed method outperform CRBO in higher-dimensional search spaces.
When comparing the BO methods with BBRL-PPO, we find all BO show higher data-efficiency than BBRL-PPO for all cases in our experiments.

\begin{figure}[!t]
    \centering
    \includegraphics[width=3.5in]{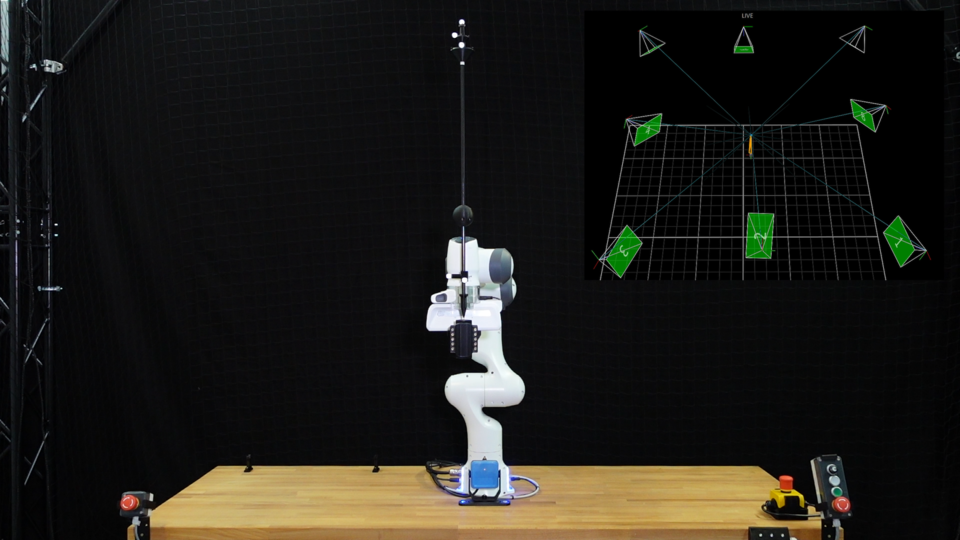}
    \caption{\textbf{Experiment setup:} The Franka Emika robot is holding a planar pendulum, and the Vicon optical motion capture tracks the pendulum object.}
    \label{fig:setup}
\end{figure}
\begin{figure}[!t]
    \centering
    \includegraphics[width=3.5in]{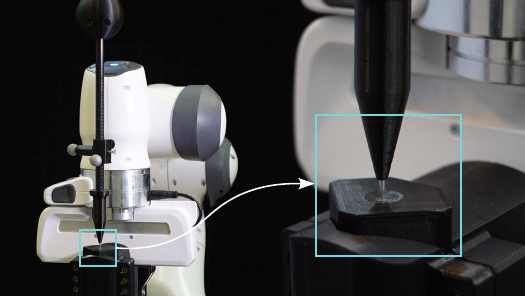}
    \caption{\textbf{Pendulum:} The pendulum is a stick with a needle, and the end-effector can move in all directions in the horizontal plane.}
    \label{fig:needle}
\end{figure}
\subsection{Learning from Scratch: Impromptu Trajectory Tracking}
Trajectory tracking control is a fundamental and essential aspect of robotic systems, yet designing and fine-tuning classical controllers can be rather labor-intensive with hidden dynamics, making it challenging to achieve optimal tracking precision~\cite{li2017deep}. 
In this experiment, we investigate how our proposed methods learn to improve the tracking performance of such robotic systems.
We choose a robotic pole-balancing setting, where the robotic arm has two different level control tasks: \emph{(i)} to ensure that the pole does not fall over (low level), and \emph{(ii)} to get the tip of the pole to precisely track a preset trajectory (task level).
The predefined trajectory is only encoded in a performance function at the task level, and the low level pendulum balancing control has no prior knowledge of the predefined trajectory.
The main motivation for robotic pole-balancing setting is that it emulates universal robot trajectory tracking in an abstract and fairly easy way, since sophisticated tasks are usually built upon low level robot kinematic control. Further, the problem of planning, in particular with unknown nonlinear dynamics, at the task level can be treated as a search problem. The optimizer operates at the task level, and learns to improve the tracking performance. 

\paragraph{Robot test bed setup}
The test bed setup, shown in Fig.~\ref{fig:setup}, consists of a Franka Emika Robot, a planar pendulum, and a Vicon optical motion capture system. The Franka Emika robot is a seven-joint open-chain manipulator that serves as a cart in a classic pendulum balancing problem. The current state of the robot and the torque control command are transmitted via the Franka Control Interface at \SI{1}{\kilo\hertz}. For our experiment, we published an open-source ROS package\footnote{The code of pendulum balancing is available at \href{https://github.com/Data-Science-in-Mechanical-Engineering/franka_pendulum}{https://github.com/Data-Science-in-Mechanical-Engineering/franka\_pendulum}.} that enables simultaneous simulation and hardware control of an inverted pendulum on top of the native interface \emph{libfranka}.

The pendulum is a stick with a needle to yield low-friction contact to the end-effector. The end-effector can move in all directions in the horizontal plane (see \fig\ref{fig:needle}).
% We extended the degrees of freedom of the pole to induce more dynamics and make the problem setting more challenging. Instead of attaching a joint that only rotates around one axis to the robot, a needle functioning as two stacked orthogonal revolute joints of the pole is placed in the center of the base held by the gripper; therefore, the gripper of the robot shall move in a horizontal plane to stabilize the planar pendulum.
%
The Vicon motion capture system evaluates the position of special spherical markers and considers an object as an array of marker coordinates in the object reference frame. A transformation from the object coordinate to the world coordinate is applied so that we obtain measurements of the pole angle and estimates of the angular velocity by numerical differentiation at \SI{100}{\hertz}.

Similar to \cite{schaal1996learning, marco2016automatic}, the dynamics of the balancing problem can be described as:
\begin{equation}
  \begin{aligned}
    & ml^2 I     \begin{bmatrix}
      \ddot \phi \\
      \ddot \theta
    \end{bmatrix} - mgl I
    \begin{bmatrix}
      \sin \phi \\
      \sin \theta
    \end{bmatrix} + mgl I
    \begin{bmatrix}
      \cos \phi u_x\\
      \cos \theta u_y
    \end{bmatrix} + \xi \begin{bmatrix}
      \dot \phi \\
      \dot \theta
    \end{bmatrix} = 0, \\
    & \begin{bmatrix}
      \ddot x \\
      \ddot y
    \end{bmatrix} = \begin{bmatrix}
    u_x \\ u_y
    \end{bmatrix} + \begin{bmatrix}
      a_x \\ a_y
    \end{bmatrix},
  \end{aligned} \label{eq:pole}
\end{equation}
where \(I\) is an identity matrix, \(\bm q = [\phi, \theta]\tran\) is the pole angle vector from the vertical  line in two directions, \(\bm u = [u_x, u_y]\tran\) is the end-effector acceleration vector, and \(\bm \eta = [x, y]\tran\) is the distance vector of the base from its origin \((x_0, y_0)=(\SI{0.5}{\m},\SI{0}{\m})\). The pole mass \(m\) is \SI{0.1593}{\kg}, and the center of mass \(l\) lies at \SI{0.463}{\m}.

The applied control law to the pendulum can be written as 
\begin{equation} \label{eq:pole_controller}
  \bm u = F\bm x + \bm a.
\end{equation}
In \eqref{eq:pole_controller}, $F\bm x$ is state-feedback control that stabilizes the pole at the origin, using a well-established discrete-time Linear Quadratic Regulator (LQR),
where \(\bm x = [\phi, \dot \phi, x, \dot x, \theta, \dot \theta, y, \dot y]\tran\) is the full state vector of the balancing system. 
The control gain \(F\) is computed by solving the LQR problem to minimize
\begin{equation}
    f_{\rm LQR} = \int_{0}^{\infty} \left(x\tran Q x + u\tran R u\right){\rm d}t,
\end{equation}
where \(Q = 10 I, R = I\). Following this, \(F = [\bm F_v, \bm 0; \bm 0, \bm F_v]\), \(\bm F_v = [45.4, 11.9, 3.16, 5.74]\).

Further, in \eqref{eq:pole_controller} we use DMPs to generate the extra acceleration $\bm a = [a_x, a_y]\tran$ to the stabilized inverted pole.
A rhythmic DMP as proposed in \cite{ijspeert2013dynamical} has been adopted to generate the periodic acceleration.
The external acceleration \(a_x\) is
\begin{equation}
    a_x = \frac{\sum^N_{i=1}\Phi_i(t) w_i}{\sum^N_{i=1}\Phi_i(t)} r,
\end{equation}
where \(\Phi_i(t) = \exp(h_i(\cos(\omega t - c_i)-1))\) are von Mises basis functions that are centered at constants \(c_i\), \(r\) is the amplitude, and \(w_i\) are adjustable weights. 
In the experiment, we set \(r = 1, h = 1\), and the \(c_i\) are evenly placed in \([0, T_s)\). We adopt the same DMP with twelve basis functions for \(a_x\) and \(a_y\) so that the planar policy is parameterized by 24 weights \(w_i\).
The initial value of each weight is 0, \ie $a = 0$, and the robot will hold the pole at its origin during the first rollout. 

The control input \(\bm u\), also the reference acceleration of the gripper, in (\ref{eq:pole_controller}) is then generated by the end-effector of the Franka Emika robot, realized by Cartesian space controller.

By learning parameters of the DMPs, the optimizer tries to make the tip of the pole follow a reference trajectory. 
We define the reference trajectory as a Lissajous curve:
\begin{equation}
\begin{aligned}
x_d &= 0.2 \sin(\omega t)\cos(\omega t) + 0.5, \\
y_d &= 0.12 \sin(\omega t), 
\end{aligned}
\end{equation}
where \(t \in [0, T_s),\omega = 2\pi / T_s\). 

For each rollout, the robot tracks the trajectory for a fixed time \(T_s = \) \SI{20}{\second}. In this way, the end-effector will move back to the origin at the end of each trial, saving the time for initializing the test bed before each rollout.

The performance of a rollout on the robot is defined by the cost function
\begin{equation}
    f = \int_0^{T_s} \left(1.2 \|x-x_{\mathrm{d}}\|_2 + \|y - y_{\mathrm{d}}\|_2\right){\rm d}t. \label{eq:reward}
\end{equation}
We discretized the robot-pendulum tracking system using zero-order hold with a frequency of \SI{100}{\hertz}.
As such each rollout contains 2000 time steps.
\(f_{\simu}\) is calculated using the same cost function from the simulator.
Norms are weighted to encourage simultaneous growth in both directions. 

\paragraph{Optimizer setup} We adopt three different optimizer settings in the experiment: \emph{(i)} HCI-GIBO on the real robot, \emph{(ii)} S-HCI-GIBO on the real robot with the help of a simulator, and \emph{(iii)} HCI-GIBO on the simulator (sim only). For the setting \emph{(iii)}, the final policy learned on the simulator is evaluated on the real robot. We perform five consecutive trials for each setting to report average performance and the deviation. These settings will help us confirm whether the simulator can improve data efficiency by comparing \emph{(i)} and \emph{(ii)}, and whether the SimToReal is vital to achieve the learning goal by comparing \emph{(ii)} and \emph{(iii)}.

For the HCI-GIBO with no simulator, we use the SE kernel
\begin{equation}
K(\theta_1, \theta_2)= \nu^2 \exp \left(-\|\theta_1-\theta_2\|^2/(2\lambda) \right),
\end{equation}
with hand-tuned variance $\nu=2$ and lengthscale $\lambda=0.3$ for all 24 dimensions. For the S-HCI-GIBO, we use the kernel introduced in \eqref{eq:new_kernel}. Note that in \eqref{eq:new_kernel}, $K_f$ and $K_m$ are both SE kernels, and in our experiment, the variance and lengthscale for $K_f$ are the same as the HCI-GIBO setting. For $K_m$, we select $\nu=0.5$, $\lambda=0.35$.

The improvement confidence $\alpha$ and the stepsize $\eta$ for all optimizers is 0.95 and 0.2, respectively. The SimToReal switching threshold $\beta$ for S-HCI-GIBO is $1$.
\begin{figure}[!t]
    \centering
    \includegraphics[width=3.5in]{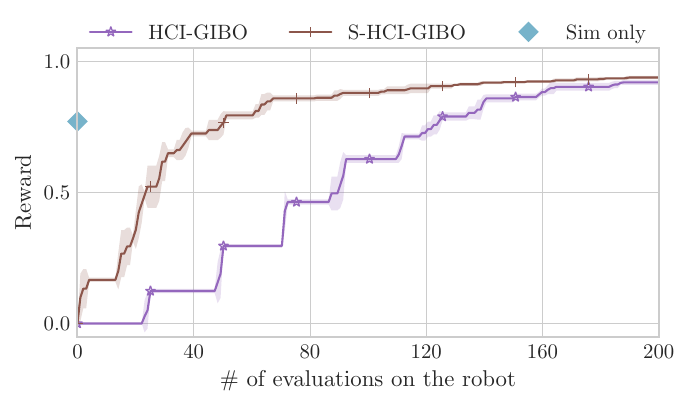}
    \caption{\textbf{Progress of HCI-GIBO and S-HCI-GIBO for the robot experiment:} Averaged over five trials (solid lines). We also run the HCI-GIBO on the simulator, which is denoted by the light-blue dot (``sim only''), and then deploy the learned policy on the real robot. Transferring the policy learned on the simulator cannot achieve fine performance. The shaded regions reported on the x-axis show the standard deviation. }
    \label{fig:experiment_reward}
\end{figure}

\begin{table}[!t]
\centering
\caption{Average evaluations on the robot for a reward level.}
\label{tab:experiment}
\begin{tabular}{@{}cccccccccc@{}}
\toprule
        & \multicolumn{9}{c}{Average evaluations to reach reward level}               \\ \cmidrule(l){2-10} 
Reward level        & 0.1 & 0.2 & 0.3 & 0.4 & 0.5 & 0.6 & 0.7 & 0.8 & 0.9 \\ \midrule
HCI-GIBO     & 24  & 49  & 62  & 71  & 90  & 90  & 113 & 132 & 175 \\ \cmidrule(r){1-1}
S-HCI-GIBO & 2   & 16  & 16  & 21  & 26  & 28  & 37  & 58  & 117 \\ \bottomrule
\end{tabular}
\end{table}
\begin{figure}[!t]
    \centering
    \includegraphics[width=3.5in]{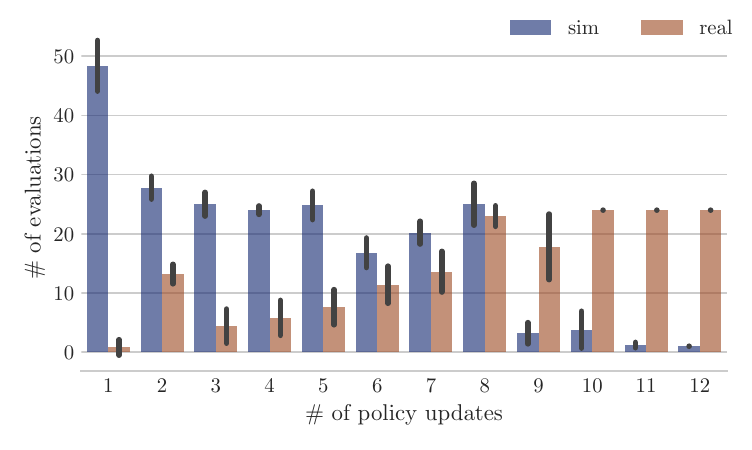}
    \caption{\textbf{S-HCI-GIBO queries on sim and real:} Averaged over five trials (bars). The vertical black lines show the standard deviation.}
    \label{fig:trade_off}
\end{figure}

\paragraph{Results}
\fig\ref{fig:experiment_reward} shows the reward with respect to queries executed on the real robot. Since the robot learns the policy from scratch and the algorithm improves the performance after each policy update, we define the performance of the initial policy as the ``worst'' performance while the ``best'' performance based on \eqref{eq:reward} is 0. We use the possible maximum reward increase in \fig\ref{fig:experiment_reward} so that it is consistent with the results from the synthetic experiments. 

The training curves indicate that S-HCI-GIBO requires less than half of the queries from the real robot to reach $80\%$ of the maximal reward. More details are provided in \tabl\ref{tab:experiment}. To further investigate the significance of sim-to-real transfer, we ran the HCI-GIBO algorithm with 200 evaluations on the simulator, transferred the optimal learned policy to the robot, and then measured the performance in the real world. \fig\ref{fig:experiment_reward} clearly shows the limitation of the simulator: even though we use a sophisticated simulator, a policy adequate to solve the task is hard to obtain by relying on the simulator alone.

We show the trade-off between queries from the simulator and the robot in \fig\ref{fig:trade_off}. In this figure, we observe that, as expected, S-HCI-GIBO appears to query \(f_{\simu}\) more frequently than \(f\) at the beginning of the learning process. This makes sense as it is designed to start with the simulator and exhaust the potential information from this cheaper source. Interestingly, after eight policy updates, we observe a steep decrease in the number of queries from the simulator. We attribute this to the much shallower gradients where additional data from the simulator is not able to reduce gradient uncertainty.
\begin{figure}[!t]
    \centering
    \includegraphics[width=3.5in]{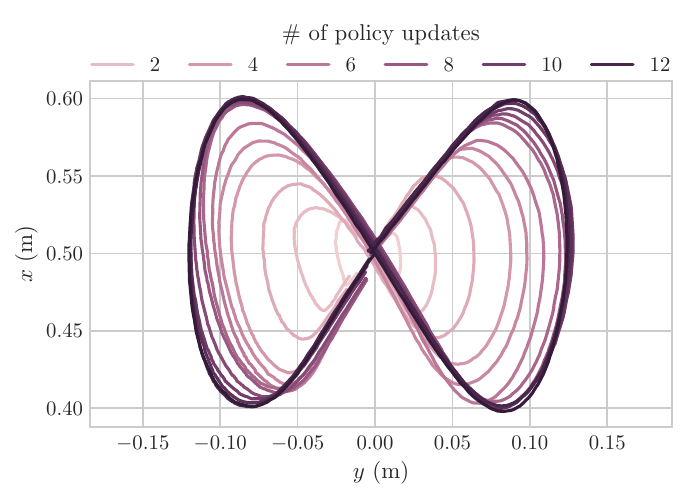}
    \caption{\textbf{Improvement:} The trajectory of the pole tip after each gradient step converges towards the desired one.}
    \label{fig:trajectory}
\end{figure}
The trajectory of the pole tip after each gradient step is shown in \fig\ref{fig:trajectory}.
The trajectory becomes closer to its reference in each policy update.

\section{Conclusion}
In this work, we introduce a novel zero-order optimizer (HCI-GIBO) and an extension to incorporate additional data from simulations (S-HCI-GIBO). 
Both algorithms improve the sample efficiency for black-box robot learning tasks and overcome difficulties in solving high dimensional problems by BO, a popular controller tuning method. 
The HCI-GIBO method employs an active sampling strategy that reduces the uncertainties of gradient estimates, quantifies the improvement confidence to adaptively control the required queries for policy updates, and leverages the information from a simulator. 
Our large-scale synthetic experiment demonstrates that the proposed algorithms outperform the baseline algorithms, especially in higher dimensions.
The robot experiment results show that these algorithms are applicable to real-world tasks.

\paragraph*{Limitations} Our proposed methods rely on local optimization, and inherit their local convergence. As such, the learning outcome depends on the initial parameters and it may be difficult to learn complex behavior with uniformed initialization. The proposed methods work best when they are used to fine-tune policies, e.g., initialized by RL or a human expert.
Another potential drawback of our approach are Lipschitz assumptions required for the theoretical results. However, there are some preliminary results that show GPs are also able to estimate sub-gradients \cite{wu2023behavior}.
More generally, the data-efficiency of BO methods is also a function of the prior knowledge of the objective; only informed priors lead to the high data-efficiency required when learning on hardware.

\paragraph*{Future work} A direction for future research is to extend the convergence proof for GIBO \cite{wu2023behavior} to the multi-fidelity setting.
Currently our proposed methods are limited to use only one simulator, and we are working on a scheme that supports more information sources.
Given that safety in robot learning is crucial, we plan to investigate how to extend the proposed methods for constrained optimization problems. This could be achieved by using Lagrangian relaxation.
Further, to evaluate the efficacy of S-HCI-GIBO, we are interested in applying the proposed methods to different robot learning problems.

\section*{Acknowledgment}
The authors thank K. Sovailo, S. Giedyk, and F. Grimminger for their support with the design and setup of the experiment.

% \subfile{sections/appendix}

\printbibliography
\balance

\end{document}